\pdfoutput=1
\PassOptionsToPackage{table,HTML}{xcolor}
\documentclass[11pt]{article}

\usepackage{ACL2023}

\usepackage{times}
\usepackage{latexsym}

\usepackage[T1]{fontenc}

\usepackage[utf8]{inputenc}
\usepackage{multirow}
\usepackage{graphicx}
\usepackage{booktabs}
\usepackage{amsmath}
\usepackage{subfigure}
\usepackage{diagbox}
\usepackage{enumitem}

\usepackage{microtype}

\usepackage{inconsolata}
\usepackage{titlesec}
\usepackage{bbm}
\usepackage[breaklinks,colorlinks,linkcolor=blue]{}

\title{Towards Fully Exploiting LLM Internal States to Enhance Knowledge Boundary Perception}

\author{Shiyu Ni\textsuperscript{\rm{1,2,3}}\space\space
Keping Bi\textsuperscript{\rm{1,2,3}}\thanks{~~Corresponding authors}\space\space
Jiafeng Guo\textsuperscript{\rm{1,2,3}}\footnotemark[1]\space\space
Lulu Yu\textsuperscript{\rm{1,2,3}}\space\space
Baolong Bi\textsuperscript{\rm{1,2,3}}\space\space
Xueqi Cheng\textsuperscript{\rm{1,2,3}}\\
    \textsuperscript{\rm 1}CAS Key Lab of Network Data Science and Technology, ICT, CAS\\
    \textsuperscript{\rm 2}State Key Laboratory of AI Safety\\
    \textsuperscript{\rm 3}University of Chinese Academy of Sciences\\
    \{nishiyu23z, bikeping, guojiafeng, yululu23s, bibaolong23z, cxq\}@ict.ac.cn
}

\begin{document}
\maketitle
\begin{abstract}


Large language models (LLMs) exhibit impressive performance across diverse tasks but often struggle to accurately gauge their knowledge boundaries, leading to confident yet incorrect responses. This paper explores leveraging LLMs' internal states to enhance their perception of knowledge boundaries from efficiency and risk perspectives. We investigate whether LLMs can estimate their confidence using internal states before response generation, potentially saving computational resources. Our experiments on datasets like Natural Questions, HotpotQA, and MMLU reveal that LLMs demonstrate significant pre-generation perception, which is further refined post-generation, with perception gaps remaining stable across varying conditions. To mitigate risks in critical domains, we introduce Confidence Consistency-based Calibration ($C^3$), which assesses confidence consistency through question reformulation. $C^3$ significantly improves LLMs' ability to recognize their knowledge gaps, enhancing the unknown perception rate by 5.6\% on NQ and 4.9\% on HotpotQA. Our findings suggest that pre-generation confidence estimation can optimize efficiency, while $C^3$ effectively controls output risks, advancing the reliability of LLMs in practical applications\footnote{The code can be found at \href{https://github.com/Trustworthy-Information-Access/LLM-Knowledge-Boundary-Perception-via-Internal-States}{GitHub Repository}.}.

\end{abstract}

\section{Introduction}
Large language models (LLMs) store vast amounts of knowledge in their parameters and have demonstrated remarkable performance across various tasks~\citep{touvron2023llama, achiam2023gpt, yang2024qwen2}. However, they may hallucinate, generating responses that appear to be fluent but are factually incorrect. A reliable model should perceive its knowledge boundaries well, providing correct answers to the questions it knows and declining to answer those it does not. This requires the model to align its confidence with its actual abilities. \looseness=-1

Current research on a model's perception of its knowledge boundaries mainly involves two types of confidence: probabilistic confidence where they use the probability of generated tokens as the model’s confidence in the answer~\citep{guo2017calibration,desai2020calibration,jiang2021can,kadavath2022language,si2022prompting,kuhn2023semantic} and verbalized confidence where they teach the model to express its confidence in the answer using natural language.~\citep{lin2022teaching, yin2023large, tian2023just,xiong2023can,yang2023alignment,ni2024llms}. \citet{ni2024large} found that probabilistic confidence better reflects the model's capability than verbalized confidence.

Recent studies have demonstrated that the internal states of LLMs can indicate the factuality of texts~\citep{azaria2023internal}. Specifically, \citet{su2024unsupervised} and \citet{chen2024inside} demonstrated that LLMs' internal states can be leveraged to evaluate the factuality of self-generated content, with confidence derived from these internal states providing a more accurate reflection of the model's capabilities than probabilistic confidence. 
Building on this, this paper focuses on estimating LLMs' confidence based on their internal states, aiming to enhance their knowledge boundary perception from efficiency and risk perspectives.

On the one hand, most existing studies rely on the internal states of the model after generating a response to assess its correctness. However, this does not prevent the generation of incorrect information and introduces extra computational overhead. In contrast, humans often know whether they can answer a question simply by considering the question itself. This raises the question: is it necessary to use LLMs' internal states after generation to assess confidence? If not, the model could save computational resources by generating answers only when it is confident. 

\begin{figure}[t!]
  \centering
    \includegraphics[width=0.45\textwidth]{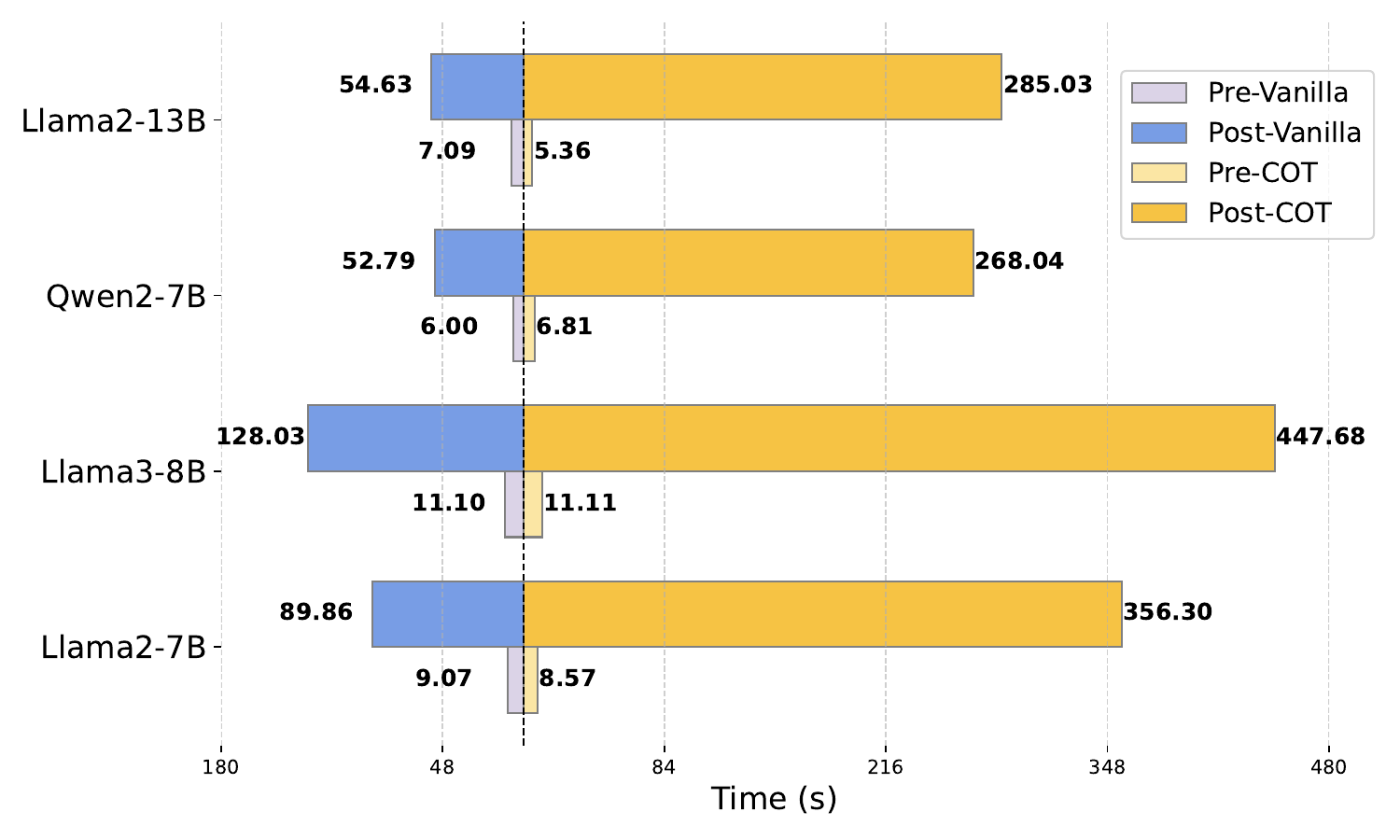}
  \caption{Time (in seconds) taken to obtain pre-generation and post-generation states for each model on the first 500 data points of the NQ test. Pre-Vanilla and Pre-COT refer to the pre-generation states obtained under the vanilla prompt and COT, respectively, while Post-Vanilla and Post-COT refer to the corresponding post-generation states.}
  \label{fig:efficiency}
  \vspace{-10pt}
\end{figure}

To explore this, we use the embeddings of the question and the full question-answer sequence to estimate the model’s perception of its knowledge boundaries before and after answer generation. We also compare the gap between these two perceptions. We conduct experiments on Natural Questions (NQ)~\citep{kwiatkowski2019natural}, HotpotQA (HQ)~\citep{yang2018hotpotqa}, and MMLU~\citep{hendrycks2020measuring} to examine the effects of question difficulty and task format. We employ Chain-of-Thought~\citep{wei2022chain} strategy to increase the information content in generated responses, aiming to explore whether this added information impacts the gap between pre- and post-generation perceptions. Furthermore, we conduct experiments with training sets of varying sizes to assess the impact of training data volume on this gap. Experimental results reveal that \textbf{LLMs exhibit a good level of pre-generation perception, and their post-generation perception will be further enhanced. The gap between these two perceptions remains relatively stable across different question difficulties, task formats, amounts of generated content, and training set sizes}. Therefore, in efficiency-critical scenarios, pre-perception can be used to determine whether generation is necessary, offering a more efficient alternative to post-generation assessment, particularly when generating lengthy content. The time for obtaining internal states before and after response generation can be found in Figure~\ref{fig:efficiency}.

On the other hand, in addition to efficiency, controlling the risk associated with model outputs is also crucial, especially in safety-critical domains like healthcare, which helps us decide when to trust the LLM. This requires accurately detecting what LLMs do not know.

To enhance LLMs' perception of what they do not know, we introduce $C^3$ (\textbf{C}onfidence \textbf{C}onsistency-based \textbf{C}alibration), inspired by human behavior, where repeated probing is used to detect inconsistencies and potential deception. $C^3$ leverages confidence consistency: if a model is confident in answering a question but loses confidence when the question format changes, this inconsistency signals potential uncertainty, indicating that the model may be overconfident.

$C^3$ has two phases: Question Reformulation and Confidence Calibration, as illustrated in Figure~\ref{fig:C3 workflow}. In the first phase, to avoid relying on additional information and to obtain questions of varying difficulty, the model is asked to generate 10 potential answers for a given question. These answers are then used to create multiple-choice questions with different numbers of answer options. Next, we calibrate the original confidence based on the consistency between the model's confidence on the original question and its confidence on each multi-choice question. Results show that \textbf{$C^3$ substantially enhances LLMs' perception of what they do not know, improving the unknown perception rate by 5.6\% on NQ and 4.9\% on HotpotQA compared to directly estimating confidence based on the original question-answer sequence}.

\section{Related Work}
Current research on how to express LLMs' confidence can be mainly divided into three categories:
\paragraph{Probabilistic Confidence.}
This series of work uses the generation probability of the answer as the model's confidence.~\citep{guo2017calibration,desai2020calibration,jiang2021can,kadavath2022language,si2022prompting,kuhn2023semantic}. \citet{guo2017calibration} found that early neural networks tend to be overconfident and mitigated this by adjusting the temperature during the generation process. Subsequently, \citet{desai2020calibration} found that pre-trained Bert-style models have a relatively clear perception of their knowledge boundaries and~\citet{jiang2021can} showed that the issue of overconfidence still persists in pre-trained language models. More recent studies have explored LLMs’ perception of their knowledge boundaries. \citet{kadavath2022language} and~\citet{si2022prompting} demonstrated LLMs can be reliable under approprite prompts. \citet{kuhn2023semantic} argued that the probability of generated tokens does not accurately reflect the probability of the generated answer and estimated LLMs' confidence in their answers based on the semantic consistency across multiple generations.

\paragraph{Verbalized Confidence.} 
With the development of LLMs, some studies have shown that LLMs can express their confidence in answers in natural language~\citep{lin2022teaching, yin2023large, tian2023just,xiong2023can,yang2023alignment,ni2024llms}. \citet{lin2022teaching} were the first to demonstrate that a model (GPT-3) can learn to express confidence about its answers using natural language. Recently, \citet{yin2023large} found that LLMs have difficulty in perceiving their knowledge boundaries and tend to be overconfident and \citet{xiong2023can} explored various methods of confidence extraction. To enhance LLMs' perception level, \citet{tian2023just} and~\citet{ni2024llms} focused on prompting methods while~\citet{yang2023alignment} proposed training methods.

\paragraph{Confidence Estimation via Internal States.}
LLMs' internal states have been found to be effective in indicating the factuality of texts.~\citep{azaria2023internal,slobodkin2023curious} and~\citep{su2024unsupervised,chen2024inside} extended this approach to detect the factuality of model’s self-generated content. This line of work utilized a shallow network (i.e., MLP) to extract confidence from the hidden states of LLMs. Compared to prob-based methods, this tends to be more accurate because converting hidden states into token probabilities results in information loss. Additionally, compared to training LLMs to express better verbalized confidence it is much more cost-efficient.

 In this paper, we exploit LLMs' internal states to enhance their knowledge boundary perception from efficiency and risk mitigation perspectives.
 
\section{Estimating LLM Confidence with Internal States}
In this section, we introduce the task formulation, how we extract internal states, and the confidence estimator.


\subsection{Task Formulation}
We introduce the task formulation of confidence estimation via LLMs' internal states here.
The process of estimating a model's confidence based on its internal states is as follows.
For a given model $ M $ and a question $ q $, it generates a response $ a_{M,q} $ and produces internal states $ I_{M,q(a)} $. Specifically, $ I_{M,q} $ refers to the internal state containing only information about the question, while $ I_{M,qa} $ represents the internal state containing information about the entire question-answer sequence:
\begin{equation}
    I_{M,q(a)}, a_{M,q}=f_M(q),
\end{equation}
 then, we estimate the model's confidence from its internal state $I_{M,q(a)}$:
\begin{equation}
    c_{M,q} = \mathcal{E}(I_{M,q(a)}),
\end{equation}
where $\mathcal{E}$ is the confidence estimator. $c_{M,q}=1$ indicates the model is confident that it knows the correct answer while $c_{M,q}=0$ means the opposite. 
The confidence estimator $\mathcal{E}$ can be learned through a dataset $D_M^{train}=\{(I_{M,q_i(a_i)}^{train}, c_{M,q_i}^{train})_{i=1}^N\}$ where $N$ is the count of samples in this dataset. The ground-truth confidence $c_{M,q_i}^{train}$ is set to 1 if the model can correctly answer the question $q_i$ (i.e., $c_{M,q_i}=1$); otherwise, it is set to 0. 
Once $\mathcal{E}$ is learned, we can perform confidence estimation during inference.

Recent works~\citep{azaria2023internal, chen2024inside, su2024unsupervised} commonly use the embedding of the last token in the generated answer to estimate the model's confidence. This state contains information from the entire question-answer sequence, potentially leading to more accurate judgments. However, relying on post-generation states does not prevent the generation of incorrect information, which can mislead users and introduce extra overhead. Therefore, in this paper, we extract representations prior to answer generation to investigate whether LLMs can sense their knowledge boundaries before response generation. The specific extraction method is detailed in Section~\S~\ref{sec:training data construction}.


\subsection{Internal States Extraction \label{sec:training data construction}}
\begin{figure}[t!]
  \centering
    \includegraphics[width=0.5\textwidth]{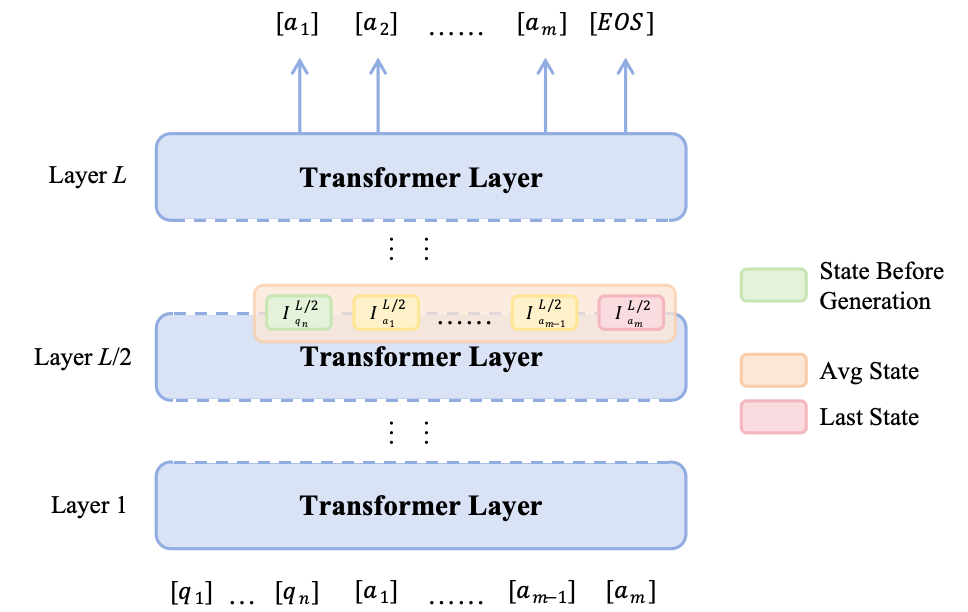}
  \caption{Internal states extraction during generation.}
  \label{fig:pooling methods}
\end{figure}

In Transformer-based models, the model performs next token prediction, where the generation of each token is based on the semantic vectors (i.e., internal states) of the preceding tokens in its sequence. For a question $q$, let the input tokens be $\{q_1, q_2, \dots, q_n\}$ and the output answer tokens $a$ be $\{a_1, a_2,\dots, a_m\}$ where $n$ and $m$ represent the count of tokens in the question and the generated answer, respectively. The internal states corresponding to the generation of each token in the answer (See Figure~\ref{fig:pooling methods}) are represented as follows:
\begin{equation}
    \{I_{q_n}^l, I_{a_1}^l, I_{a_2}^l, \dots, I_{a_{m}}^l\}_{l=1}^L,
\end{equation}
 where $I_x^l$ denotes the semantic representation of the tokens up until $x$ at layer $l$, and $L$ is the total number of layers in the model. Note that $I_{q_n}^l$ contains only information about the question.
 
\paragraph{Layer Selection}
Previous work~\citep{azaria2023internal} has found that the representations from the intermediate layers best capture the model’s awareness of factuality. Therefore, we extract representations from the intermediate layers (i.e., 16 for Llama2-Chat-7B) to construct the internal states $I$.

To investigate whether it is necessary to extract LLMs' internal states after response generation, we construct $I$ (see Figure~\ref{fig:pooling methods}) at two stages: before and after response generation. 

\paragraph{Extraction Before Response Generation.}
\begin{enumerate}
    \item \textbf{Pre-State}. We extract the state of the last token of the question $I_{q_n}^{mid}$ as $I$.
\end{enumerate}
 
\paragraph{Extraction After Response Generation.}
Unlike pre-generation states, we extract $I$ in two ways:
\begin{enumerate}
    \item \textbf{Last State}. We take the embedding of the last token from the generated answer $I_{a_{m}}^{mid}$ as $I$.
    \item \textbf{Avg State}. We take the average of the representations of each generated token in the answer $\frac{1}{m}\sum_{i=1}^{m}I_{a_i}^{mid}$ as $I$.
\end{enumerate}

\paragraph{Training Data Construction.} For each question $q$ in the training set, we prompt the model to generate a response and construct $\{I, c\}$, where $c = 1$ if the ground-truth answer is included in the response, and $c = 0$ otherwise.

\subsection{Binary Confidence Estimator}
Similar to previous works~\citep{azaria2023internal, chen2024inside, su2024unsupervised}, we take a lightweight MLP (Multi-layer Perceptron) network as the estimator to perform binary classification on the model's confidence. The estimator takes the internal states which are constructed as described in Section ~\S~\ref{sec:training data construction} as input and outputs a binary confidence label indicating whether the model is confident to provide a correct answer. This process can be mathematically expressed as:
\begin{equation}
    P(\hat{c}=1) = \sigma\left( \text{MLP} \left( I  \right) \right),
\end{equation}
where $\hat{c}$ is the predicted confidence, $\sigma$ is the sigmoid function, and $I \in \mathcal{R}^{1 \times h}$ is the internal state vector where $h$ refers to the model's hidden dimension. $W_{i}\in \mathcal{R}^{d_i \times d_{i-1}}$ where $d_i$ denotes the number of hidden units in the $i_{th}$ hidden layer (i.e., $d_0$ = $h$) and $b \in \mathcal{R}^{d_i}$ represent weights and the biases of MLP, respectively. We use a 4-layer MLP for binary classification on $I$, with the following number of hidden units in each layer: $(512 \to 64 \to 32 \to 2)$, and ReLU as the activation function. \looseness=-1

\paragraph{Training.}
We employ cross-entropy loss as the training objective:
\begin{equation}
    \mathcal{L}_{\text{CE}} = - \sum_{i=1}^N\mathbbm{1}(c_i) \log(P_i) + \mathbbm{1}(1-c_i) \log(1-P_i),
\end{equation}
where $c_i$ is the ground-truth label for the $i_{th}$ training sample and $P_i = P(\hat{c_i}=1)$.
We randomly initialize the model parameters and use the Adam optimizer with an initial learning rate of $5 \times 10^{-5}$. To enhance the reliability of the results, we train the model 30 epochs under three random seeds (0, 42, 100) and report the average performance as the final result. \looseness=-1

\paragraph{Inference.}
During inference, we can determine whether a model is confident to provide a correct answer as follows:
\begin{equation}
    \hat{c_i} = \arg\max_{y \in \{0, 1\}} P(c_i = y).
\end{equation}

\section{LLMs' Perception Before and After Response Generation}

In this section, we evaluate the gap between LLMs' perception level before and after response generation, as well as the impact of question difficulty, question format, the amount of generated content, and training data amount.

\subsection{Experimental Setup \label{sec:experimental setup}}

\paragraph{Datasets.\label{para:datasets}} We take three representative open-domain QA benchmark datasets, including Natural Questions (NQ)~\citep{kwiatkowski2019natural}, HotpotQA (HQ)~\citep{yang2018hotpotqa}, and MMLU~\citep{hendrycks2020measuring}. NQ and HotpotQA are two free-form QA datasets that primarily evaluate the model's factual knowledge, with varying levels of difficulty. NQ is constructed from Google Search queries, with annotated short and long answers. HotpotQA is a dataset consisting of question-answer pairs that require multi-hop reasoning. These pairs are collected via Amazon Mechanical Turk. MMLU is a multi-choice dataset containing questions from 57 different subjects. We conduct experiments on both free-form and multi-choice QA datasets to investigate the influence of task format. Due to space limitation, Count of samples for each dataset can be found in Table~\ref{tab:data_count} in Appendix and details of data selection is shown in Appendix~\S~\ref{sec:data selection}. For training, we randomly select 1,000 positive and 1,000 negative samples from the training set shown in Table~\ref{tab:data_count} to mitigate the impact of label imbalance. The choice of 1,000 is because all experiments in this paper include 1,000 positive and negative samples. Additionally, to examine the impact of training data size, we also evaluate performance using the full training set. Details on data construction and confidence estimation are provided in section~\S~\ref{sec:training data construction}.

\paragraph{LLMs.} We conduct experiments on four representative open-source models, including Llama2-7B-Chat~\cite{touvron2023llama}, Llama3-8B-Instruct~\cite{dubey2024llama}, Qwen2-7B-Instruct~\cite{yang2024qwen2}, and Llama2-13B-Chat~\cite{touvron2023llama}. We use half-precision for the 13B model. For all the models, we set the temperature to 1.0 and select the token with the highest probability at each position (i.e., greedy search). Unless otherwise specified, all the other parameters are set to their default values.

\paragraph{Metrics.}
Following previous research~\cite{ni2024llms}, we use accuracy to evaluate the QA performance considering a response correct if it contains the ground-truth answer. For the model's perception level, we use alignment, overconfidence, and conservativeness as the evaluation metrics. Alignment refers to the proportion of samples where the model's confidence matches its QA performance, serving as an indicator of the model's overall perception level. Overconfidence and conservativeness represent the proportions of samples where the model's confidence exceeds or falls below its actual capabilities, respectively, which illustrate why the model's perception level is not perfect.

\paragraph{Dimensions of Analysis.}
We hypothesize that question difficulty, question format, and the amount of training data may influence the gap between the model's pre- and post-generation perception. Therefore, we investigate along these dimensions.
The specific settings are detailed in Paragraph Datasets~\ref{para:datasets}. Additionally, to examine whether the information content in generated responses impacts the gap in LLMs' perception, we employ two strategies: Vanilla, where the model is simply asked to provide the correct answer, and Chain-of-Thought~\cite{wei2022chain} (COT), where the model first outputs its reasoning process before providing the final answer.

\begin{table*}[t!]
\centering
\small
\resizebox{0.95\linewidth}{!}{
{\begin{tabular}{cccccccccc}
    \toprule
     \multirow{2}{*}{\textbf{Datasets}} & \multirow{2}{*}{\textbf{Metrics}}  & \multicolumn{2}{c}{\textbf{Llama2-7B}} & \multicolumn{2}{c}{\textbf{Llama3-8B}} & \multicolumn{2}{c}{\textbf{Qwen2-7B}} & \multicolumn{2}{c}{\textbf{Llama2-13B}} \\
    \cmidrule(lr){3-4} \cmidrule(lr){5-6} \cmidrule(lr){7-8} \cmidrule(lr){9-10}
     &   & Vanilla & COT & Vanilla & COT & Vanilla & COT & Vanilla & COT \\
    \midrule
    \multirow{4}{*}{\textbf{NQ}} & Acc & 26.12 & 36.43 & 27.53 & 44.35 & 27.31 & 37.76 & 32.27 & 41.99 \\
    \cmidrule(lr){2-10}
                        & Align-P & 73.65 & 67.51 & 73.79 & 65.21 & \textbf{72.69} & 64.06	& 68.67 & 65.16  \\
                        & Align-L & 73.73 & 68.98	& 74.73 & 67.56 & 70.77 & 67.85	& \textbf{72.15} & 66.12  \\
                        & Align-A & \textbf{74.82} & \textbf{70.12} & \textbf{75.35} & \textbf{67.66} & 72.65 & \textbf{69.78}	& 71.20	& \textbf{67.03} \\
    \midrule
    \multirow{4}{*}{\textbf{HotpotQA}} & Acc & 19.93 & 29.55 & 21.63 & 36.79 & 24.96 & 33.34 & 23.69 & 33.10 \\
    \cmidrule(lr){2-10}
                              & Align-P & 79.69 & 74.36	& 78.58 & 74.61 & \textbf{79.34} & \textbf{76.79}	& 75.91 & 73.55 \\
                              & Align-L & 79.16 & \textbf{74.77}	& \textbf{80.77} & 74.59 & 75.13 & 75.20	& \textbf{77.66} & \textbf{74.10} \\
                              & Align-A & \textbf{79.91} & 72.71	& 80.61 & \textbf{74.67} & 77.77 & 75.33	& 76.32 & 72.57  \\
    \midrule
    \multirow{4}{*}{\textbf{MMLU}} & Acc & 42.20 & 45.51 & 62.49 & 63.77 & 68.72 & 68.63 & 50.58 & 51.18  \\
    \cmidrule(lr){2-10}
                          & Align-P & 62.86 & 63.83	& 71.95 & 68.17 & 69.33 & 68.68	& 64.88 & 64.25 \\
                          & Align-L & 68.11 & 66.55	& 72.86 & 70.43 & 70.02 & 72.66	& 67.75 & 67.96  \\
                          & Align-A & \textbf{68.71} & \textbf{67.95} & \textbf{73.98} & \textbf{71.61} & \textbf{72.57} & \textbf{72.74}	& \textbf{69.18} & \textbf{69.30}  \\
    \bottomrule
  \end{tabular}}
}
  \caption{QA performance and LLMs' perception of knowledge boundaries on the NQ, HotpotQA, and MMLU datasets with 2,000 training samples. Bold values denote the highest performance per model and dataset. Align-P, Align-L, and Align-A represent alignment scores for Pre-generation, Last, and Average States, respectively.}
  \label{tab:vanilla and cot res}
\end{table*}

\begin{table*}[h]
\centering
\resizebox{\linewidth}{!}{
{\begin{tabular}{cccccccccccc}
    \toprule
     \multirow{2}{*}{\textbf{Datasets}} & \multirow{2}{*}{\textbf{States}} & \multicolumn{5}{c}{\textbf{Vanilla}} & \multicolumn{5}{c}{\textbf{COT}} \\
     \cmidrule(lr){3-7} \cmidrule(lr){8-12}
     &   & Acc & Conf. & Align.$\uparrow$ & Overcon.$\downarrow$ & Conserv.$\downarrow$ & Acc & Conf. & Align.$\uparrow$ & Overcon.$\downarrow$ & Conserv.$\downarrow$ \\
    \midrule
    \multirow{3}{*}{\textbf{NQ}}& Pre-State & 27.53 & 17.38 & 73.79 & \textbf{8.02} & 18.18 & 44.35 & 41.12 & 65.21 & 15.78 & 19.01  \\
                       & Last State & 27.53 & 21.47 & 74.73 & 9.60 & \textbf{15.67} & 44.35 & 41.50 & 67.56 & \textbf{14.79} & 17.65 \\
                       & Avg State & 27.53 & 19.71 & \textbf{75.35} & 8.41 & 16.23 & 44.35 & 43.91 & \textbf{67.66} & 15.95 & \textbf{16.39} \\
   \midrule
   \multirow{3}{*}{\textbf{HQ}} & Pre-State & 21.63 & 26.91 & 78.58 & 13.35 & \textbf{8.08} & 36.79 & 31.29 & 74.61 & \textbf{9.95} & 15.44 \\
                       & Last State & 21.63 & 24.88 & \textbf{80.77} & 11.24 & 7.99 & 36.79 & 35.18 & 74.59 & 11.90 & 13.51 \\ 
                       & Avg State & 21.63 & 24.55 & 80.61 & \textbf{11.15} & 8.24 & 36.79 & 37.72 & \textbf{74.67} & 13.13 & \textbf{12.20} \\
   \midrule
   \multirow{3}{*}{\textbf{MMLU}}&Pre-State & 62.49 & 67.83 & 71.95 & 16.36 & \textbf{11.70} & 63.77 & 80.24 & 68.17 & 23.87 & 7.97 \\
                        & Last State& 62.49 & 63.39 & 72.86 & \textbf{13.68} & 13.46 & 63.77 & 75.33 & 70.43 & \textbf{20.28} & 9.29 \\
                        & Avg State& 62.49 & 64.95 & \textbf{73.98} & 13.90 & 12.12 & 63.77 & 76.97 & \textbf{71.61} & 20.51 & \textbf{7.88} \\
    \bottomrule
  \end{tabular}}
}
  \caption{Detailed perception results for Llama3-8B. Conf., Align., Overcon., and Conserv. stands for Confident Ratio, Alignment,  Overconfidence, and Conservativeness, respectively. Bold denotes the best scores on each dataset.}
  \label{tab:detail conf llama8b}
\end{table*}

\subsection{Results and Analysis}
The QA performance and alignment results for all the models, trained on 2,000 examples, can be found in Table~\ref{tab:vanilla and cot res}, and the detailed perception results for Llama3-8B-Instruct are shown in Table~\ref{tab:detail conf llama8b}. Detailed perception results for the other models can be found in Table~\ref{tab:detail conf llama7b}, \ref{tab:detail conf qwen2}, \ref{tab:detail conf llama13b}. We observe that: 

1) \textbf{LLMs can perceive their knowledge boundaries before generating the response, and incorporating the representations of the generated answer further enhances the perception.} In Table~\ref{tab:vanilla and cot res}, across all the models and datasets, Align-P consistently achieves high perception level. Align-L and Align-A often show improvement compared to Align-P. This provides us a trade-off between judging whether the model can provide the correct answer and the computational cost.
On one hand, if we determine that the model cannot provide the correct answer before it generates a response, we can stop the generation to save computational cost, especially when responses tend to be long. On the other hand, making the judgment after generation improves the accuracy of the assessment.



2) \textbf{Including reasoning process in the output does not widen the gap between an LLM's perception level before and after generating a response, but it may reduce the model's overall perception level.} Table~\ref{tab:vanilla and cot res} shows that COT does not markedly increase the gap between Align-P and Align-L (or Align-A) in any scenario. Additionally, in free-form QA tasks, COT often improves QA performance but tends to reduce the model's perception level.  From Table~\ref{tab:detail conf llama8b}, it can be seen that COT leads to an increase in the model's confidence. However, this change in confidence does not align with the changes in QA performance, which may increase the overconfidence or conservativeness, thereby harming the alignment.
However, on MMLU, COT has no significant impact on QA performance, similar to the previous findings~\citep{sprague2024cot}, and its effect on perception level shows no clear pattern.

\begin{figure}[t!]
  \centering
    \includegraphics[width=0.45\textwidth]{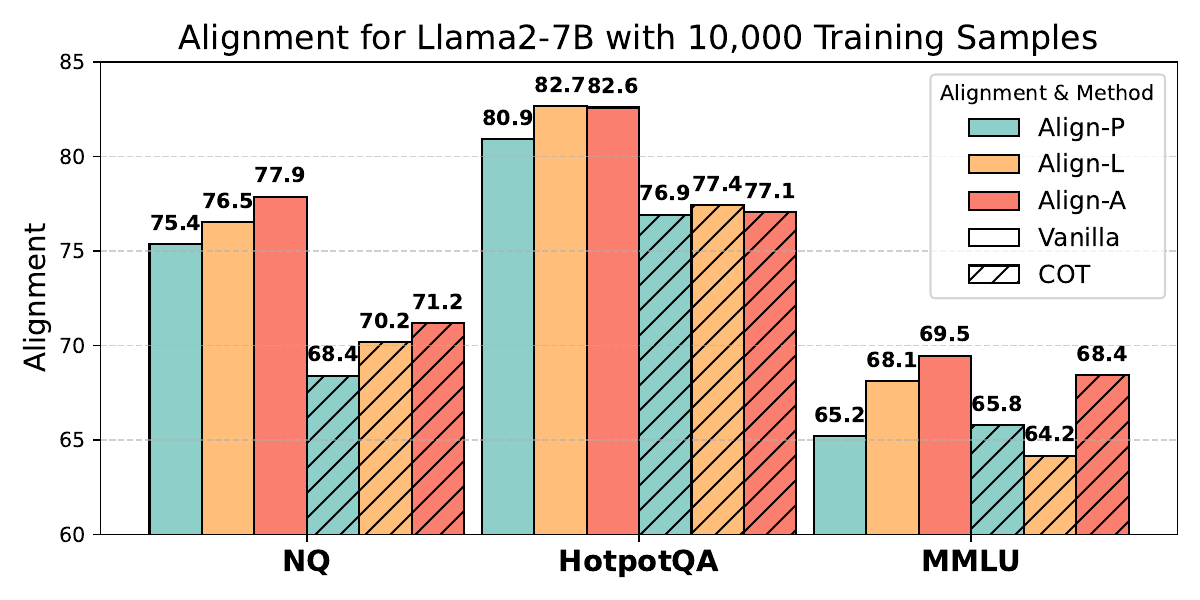}
  \caption{Llama2-7B's perception of its knowledge boundaries with 10,000 training samples.}
  \label{fig:Alignment for Llama2-7B with 1w samples}
  \vspace{-20pt}
\end{figure}

\begin{figure*}[t!]
  \centering
    \includegraphics[width=\textwidth]{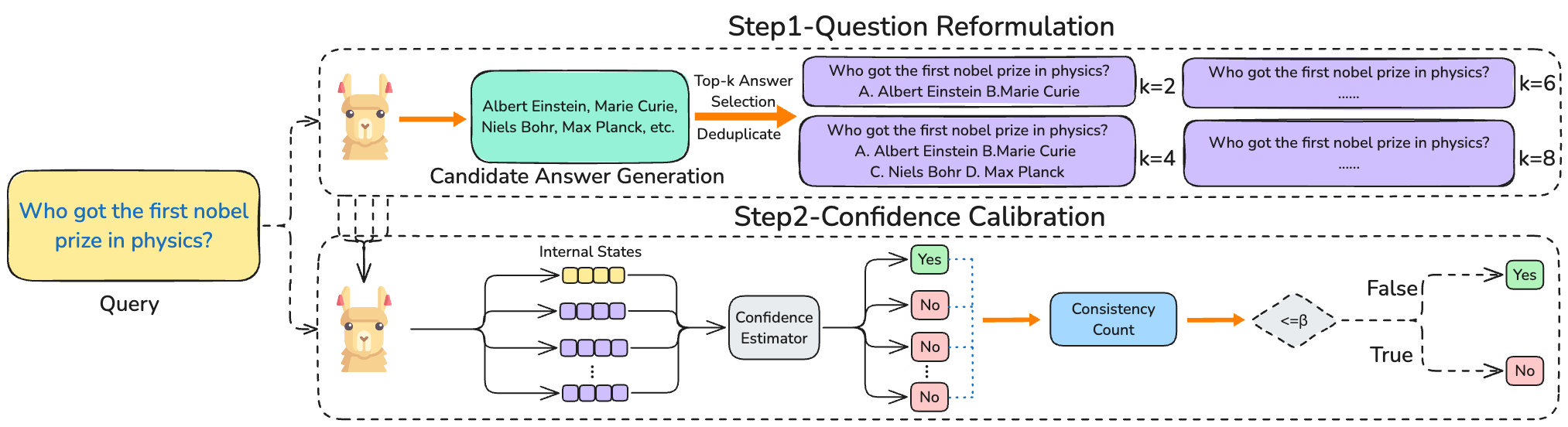}
  \caption{Workflow of $C^3$. $C^3$ includes two phases: Question Reformulation and Confidence Calibration. First, we ask the model to generate multiple answers and reformulate the original question into multiple-choice (MC) questions. Then, we estimate the model's confidence via its internal states, and calibrates its original confidence based on the consistency between its confidence in answering the original and reformatted questions.}
  \label{fig:C3 workflow}
  \vspace{-15pt}
\end{figure*}

3) \textbf{LLMs exhibit better perception of their knowledge boundaries on more difficult questions.} The results from Table~\ref{tab:vanilla and cot res} and Table~\ref{tab:detail conf llama8b} show that, compared to NQ, the model's QA performance on HotpotQA is lower, yet its confidence remains at a comparable or even higher level. This reduces conservativeness and improves alignment.

4) \textbf{The perception levels of LLMs before and after response generation can be improved with additional training data, though the gap between these levels remains nearly unchanged.} Figure~\ref{fig:Alignment for Llama2-7B with 1w samples} illustrates the perception level of Llama2-7B trained on 10,000 samples. Compared to the results in Table~\ref{tab:vanilla and cot res}, the gap between the LLMs' perception levels before and after response generation remains almost identical. A detailed analysis can be found in Appendix~\S~\ref{sec:effects of training sample amount}. \looseness=-1

\section{$C^3$: Confidence Consistency-Based Calibration}
A model introduces risks when it provides incorrect answers, which is especially unacceptable in safety-critical domains such as healthcare. In this section, we present $C^3$, a method aimed at enhancing LLMs' perception of what they do not know.

\subsection{Overview}

\begin{table*}[htbp]
\centering
\scalebox{0.9}{\begin{tabular}{cccccccccc}
    \toprule
     \multirow{2}{*}{\textbf{Models}} & \multirow{2}{*}{\textbf{Methods}} & \multicolumn{4}{c}{\textbf{NQ}} & \multicolumn{4}{c}{\textbf{HotpotQA}} \\
     \cmidrule(lr){3-6} \cmidrule(lr){7-10}
     & & \textbf{Conf.} & \textbf{UPR}$\uparrow$ & \textbf{Overcon.}$\downarrow$  & \textbf{Align.}$\uparrow$ & \textbf{Conf.} &  \textbf{UPR}$\uparrow$ & \textbf{Overcon.}$\downarrow$ & \textbf{Align.}$\uparrow$  \\
    \midrule
    \multirow{2}{*}{\textbf{Llama2-7B}} & Vanilla & 21.59 & 85.29	& 10.87 & 73.73 & 25.91 & 83.25 & 13.41 & 79.16 \\
                       &  $C^3$ & 14.23 & \textbf{91.24} & \textbf{6.47} & \textbf{75.16} & 21.56 & \textbf{86.93} & \textbf{10.46} & \textbf{80.71} \\
   \midrule
   \multirow{2}{*}{\textbf{Llama3-8B}} & Vanilla & 21.47 & 86.74	& 9.60	& 74.73 & 24.88 & 85.66 & 11.24 & 80.77 \\
                       & $C^3$ & 15.37	& \textbf{91.53} & \textbf{6.14} & \textbf{75.56} & 18.89 & \textbf{89.85} & \textbf{7.95} & \textbf{81.35} \\
   \midrule
   \multirow{2}{*}{\textbf{Qwen2-7B}}  & Vanilla & 29.41 & 78.46 & 15.66 & 70.77 & 29.95 & 80.12	& 14.93 & 75.13 \\
                       & $C^3$ & 22.82 & \textbf{84.51} & \textbf{11.26} & \textbf{72.99} & 24.16 & \textbf{85.06} & \textbf{11.21} & \textbf{76.79} \\
   \midrule
   \multirow{2}{*}{\textbf{Llama2-13B}}  & Vanilla & 26.92 & 83.39	& 11.25 & 72.15 & 25.10 & 84.45 & 11.88 & 77.66 \\
                       & $C^3$ & 17.79	& \textbf{90.32} & \textbf{6.56} & \textbf{72.41} & 20.47 & \textbf{88.40} & \textbf{8.85} & \textbf{79.08}\\
    \bottomrule
  \end{tabular}}
  \caption{The results of LLMs' perception level of their knowledge boundaries after calibration with $C^3$. Conf., Overcon., and Align. stands for Confident Ratio, Overconfidence, and Alignment, respectively. Bold indicates the best scores for each model and the results are based on Last State.}
  \label{tab:enhanced align last}
  \vspace{-10pt}
\end{table*}

\textit{If someone truly knows the correct answer, they will remain confident in their ability to answer the question correctly, even when the question is asked in different ways}. Inspired by this, we think a model may be overconfident if the model is confident in its ability to answer a question correctly but loses confidence when the question format changes. In such cases, the model's original confidence should be calibrated. This approach, which reduces overconfidence by leveraging the consistency of the model's confidence across differently phrased questions, is termed $C^3$ (\textbf{C}onfidence \textbf{C}onsistency-based \textbf{C}alibration). We focus on calibrating the model's confidence on free-form questions cause they are the most commonly used question format in practice. \looseness=-1

\subsection{Methodology}
We aim to ask the model using different question formats and leverage the consistency of the model's confidence across these formats to calibrate its confidence. Therefore, $C^3$ includes two phases: \textbf{Question Reformulation} and \textbf{Confidence Calibration}, as shown in Figure~\ref{fig:C3 workflow}.
\paragraph{Step1-Question Reformulation.} 
To avoid relying on additional information and to obtain questions of varying difficulty, we ask the model to generate multiple candidate answers and use these answers to construct multiple-choice questions.
The process can be described in the following two steps:

\begin{enumerate}
\vspace{-5pt}
    \item For question $q$, we first ask the model to generate $\alpha$ (i.e., $\alpha=10$) possible answers. Each model is able to generate more than 8 unique answers on average for questions across all the datasets, with the correctness rate of earlier-generated answers being higher. The analysis of these generated answers is provided in Appendix~\S~\ref{sec:candidate answers analysis}, and the prompt can be found in Appendix~\S~\ref{sec:prompts}.
    \vspace{-0.2cm}
    \item We deduplicate the candidate answers and take the top-$k$ (in order) as the options to construct the multi-choice question $\text{MC}_{k}$. This prevents LLMs from becoming uncertain due to the absence of the correct answer among the options. Further analysis can be found in Appendix~\ref{sec:candidate answers analysis}.
\end{enumerate}
\vspace{-5pt}
We set $k$ to 2, 4, 6, and 8, respectively.

\paragraph{Step2-Confidence Calibration} We check the consistency between the model's confidence $c_{oq}$ in the original question $q$ and its confidence $c_{mc_k}$ in the constructed multiple-choice questions $\text{MC}_k$ to calibrate $c_{oq}$. The specific strategy is as follows:

\begin{enumerate}
    \item We estimate the model's confidence for each question based on its internal states after generating the response according to Section~\S~\ref{sec:training data construction}
    \item $c_{oq}$ is refined based on multiple $c_{mc_{\alpha}}$as:
        \begin{equation}
            c_{oq} = 
            \begin{cases} 
            0, & \text{if } c_{oq} = 1 \text{and} \sum_{k \in K}^{}{c_{mc_{k}}} \leq \beta, \\
            c_{oq}, & \text{otherwise},
            \end{cases}
        \end{equation}
    where $K$ is the set of $k$ values. We set $K$ to \{2, 4, 6, 8\} and $\beta$ to 0.
\end{enumerate}

\vspace{-5pt}
\paragraph{UPR Evaluation.} To assess the ability to detect what the model does not know, we introduce \textbf{U}nknown \textbf{P}erception \textbf{R}ate (UPR). The UPR can be calculated as:
\begin{equation}
    \text{UPR} = \frac{\sum_{i=1}^n \mathbbm{1}(Acc(a_i) = 0 \text{ and } c_i = 0)}{\sum_{i=1}^n \mathbbm{1}(Acc(a_i) = 0)},
\end{equation}
The rest of the experimental settings are the same as in Section~\S\ref{sec:experimental setup}.

\vspace{-5pt}
\subsection{Results and Analysis}
The performance of $C^3$ based on Last State is presented in Table~\ref{tab:enhanced align last}. It shows that:
1) \textbf{$C^3$ substantially enhances LLMs' perception of what they do not know.} Compared to the vanilla method, $C^3$ substantially improves UPR in all the cases. This improvement occurs because the method reduces the proportion of samples where the LLMs are confident but provides wrong answers, which mitigates the LLMs' overconfidence.
2) \textbf{$C^3$ does not excessively calibrate the LLMs' confidence.} Reducing the model's confidence significantly decreases overconfidence, while slightly increasing conservativeness (See Table~\ref{tab:question reformulation performance}). Overall alignment consistently improves, suggesting that $C^3$ does not excessively calibrate the model's confidence.
The performance of $C^3$ based on Avg State shows similar conclusions and can be found in Table~\ref{tab:enhanced align avg} in Appendix. \looseness=-1

\begin{figure}[htbp]
  \centering
    \includegraphics[width=0.45\textwidth]{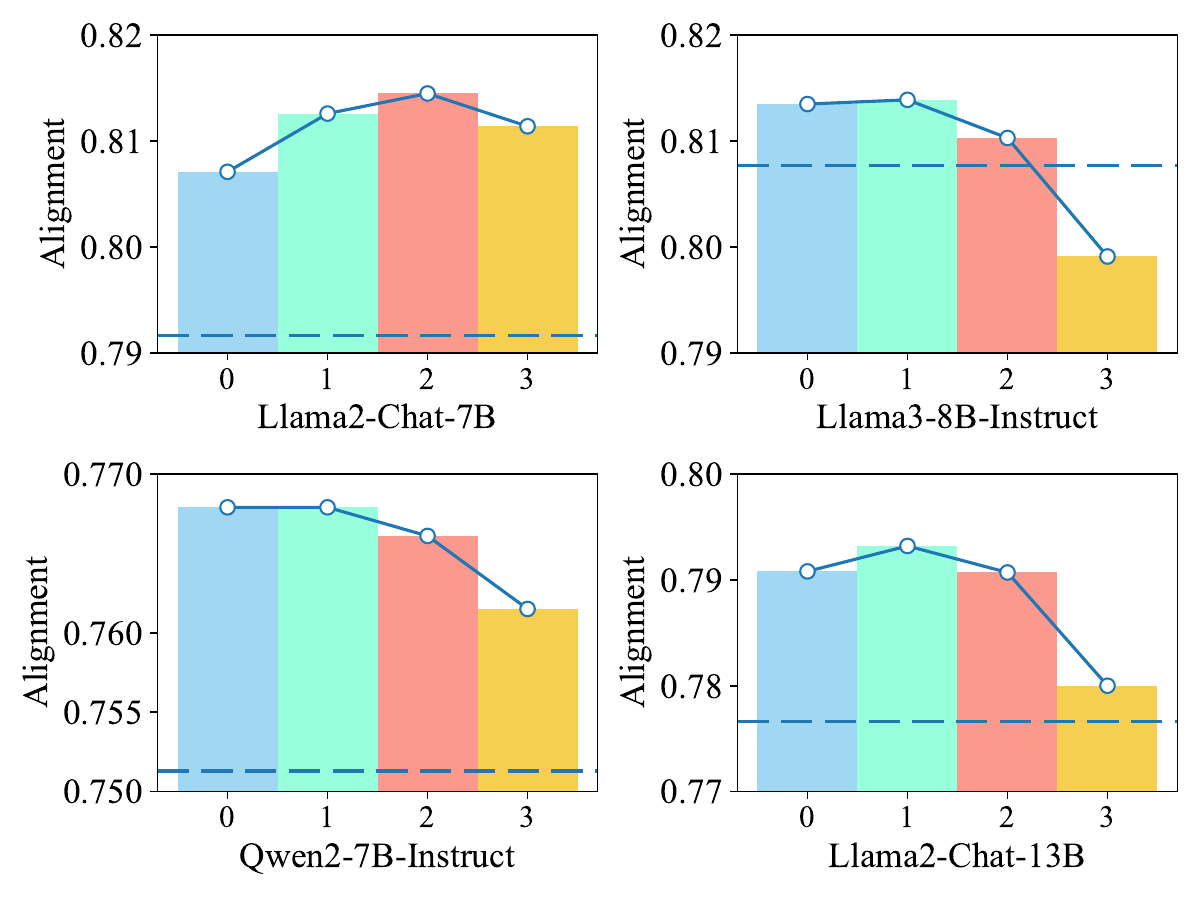}
  \caption{The alignment scores of $C^3$ under different $\beta$ when using Last State on HotpotQA. The horizontal line represents the alignment score without $C^3$.}
  \label{fig:align_hq}
  \vspace{-15pt}
\end{figure}

\paragraph{Effects of $\beta$} A larger $\beta$ results in more samples being calibrated, which aids risk mitigation but may lead to overly calibrated outcomes. The alignment score improves across almost all $\beta$ values, as shown in Figure~\ref{fig:align_hq}. This suggests that $C^3$ does not excessively adjust the model's confidence. Smaller values of $\beta$ may lead to good alignment but can limit risk mitigation, as overly strict calibration criteria may emerge. We can adjust $\beta$ to balance risk mitigation and alignment. Alignment scores under different $\beta$ values on NQ can be found in Figure~\ref{fig:align_nq} in Appendix. \looseness=-1

\vspace{-8pt}
\section{Conclusion}
\vspace{-5pt}
In this paper, we first examine LLMs' ability to assess the factuality of their responses using internal states before and after response generation. Our findings show that LLMs can predict the correctness of their answers prior to generation, providing a cost-efficient approach that avoids inference, with this ability further enhanced post-generation. 
Next, we introduce $C^3$ (Confidence Consistency-based Calibration), a method to refine the model's perception after response generation. Experimental results demonstrate that $C^3$ substantially improves LLMs' ability to recognize what they do not know and consistently enhances their overall perception.

\clearpage
\section*{Limitations}
First, we conduct research based on binary confidence to determine whether the model's output can be trusted. Further exploration is needed in the future for more fine-grained confidence. Second, we focus on the model's perception of its factual knowledge boundaries. The model's perception of its non-factual knowledge boundaries require further investigation. Third, due to resource limitations, we conduct our experiments only on 7B and 13B models. The effectiveness of our approach on larger models remains to be validated.

\section*{Ethics Statement}
We approach ethics with great care. In this paper, all the datasets and models we use are open-source. Additionally, the methods we propose aim to enhance the reliability of LLMs' responses and do not encourage or induce the model to generate any harmful information.

\section*{Acknowledgements}
This work was funded by the National Natural Science Foundation of China (NSFC) under Grants No. 62302486, the Innovation Project of ICT CAS under Grants No. E361140, the CAS Special Research Assistant Funding Project, the project under Grants No. JCKY2022130C039, the Strategic Priority Research Program of the CAS under Grants No. XDB0680102, and the NSFC Grant No. 62441229.


\bibliography{anthology,custom}

\begin{thebibliography}{27}
\expandafter\ifx\csname natexlab\endcsname\relax\def\natexlab#1{#1}\fi

\bibitem[{Achiam et~al.(2023)Achiam, Adler, Agarwal, Ahmad, Akkaya, Aleman, Almeida, Altenschmidt, Altman, Anadkat et~al.}]{achiam2023gpt}
Josh Achiam, Steven Adler, Sandhini Agarwal, Lama Ahmad, Ilge Akkaya, Florencia~Leoni Aleman, Diogo Almeida, Janko Altenschmidt, Sam Altman, Shyamal Anadkat, et~al. 2023.
\newblock Gpt-4 technical report.
\newblock \emph{arXiv preprint arXiv:2303.08774}.

\bibitem[{Azaria and Mitchell(2023)}]{azaria2023internal}
Amos Azaria and Tom Mitchell. 2023.
\newblock The internal state of an llm knows when it's lying.
\newblock \emph{arXiv preprint arXiv:2304.13734}.

\bibitem[{Chen et~al.(2024)Chen, Liu, Chen, Gu, Wu, Tao, Fu, and Ye}]{chen2024inside}
Chao Chen, Kai Liu, Ze~Chen, Yi~Gu, Yue Wu, Mingyuan Tao, Zhihang Fu, and Jieping Ye. 2024.
\newblock Inside: Llms' internal states retain the power of hallucination detection.
\newblock \emph{arXiv preprint arXiv:2402.03744}.

\bibitem[{Desai and Durrett(2020)}]{desai2020calibration}
Shrey Desai and Greg Durrett. 2020.
\newblock Calibration of pre-trained transformers.
\newblock \emph{arXiv preprint arXiv:2003.07892}.

\bibitem[{Dubey et~al.(2024)Dubey, Jauhri, Pandey, Kadian, Al-Dahle, Letman, Mathur, Schelten, Yang, Fan et~al.}]{dubey2024llama}
Abhimanyu Dubey, Abhinav Jauhri, Abhinav Pandey, Abhishek Kadian, Ahmad Al-Dahle, Aiesha Letman, Akhil Mathur, Alan Schelten, Amy Yang, Angela Fan, et~al. 2024.
\newblock The llama 3 herd of models.
\newblock \emph{arXiv preprint arXiv:2407.21783}.

\bibitem[{Guo et~al.(2017)Guo, Pleiss, Sun, and Weinberger}]{guo2017calibration}
Chuan Guo, Geoff Pleiss, Yu~Sun, and Kilian~Q Weinberger. 2017.
\newblock On calibration of modern neural networks.
\newblock In \emph{International conference on machine learning}, pages 1321--1330. PMLR.

\bibitem[{Hendrycks et~al.(2020)Hendrycks, Burns, Basart, Zou, Mazeika, Song, and Steinhardt}]{hendrycks2020measuring}
Dan Hendrycks, Collin Burns, Steven Basart, Andy Zou, Mantas Mazeika, Dawn Song, and Jacob Steinhardt. 2020.
\newblock Measuring massive multitask language understanding.
\newblock \emph{arXiv preprint arXiv:2009.03300}.

\bibitem[{Jiang et~al.(2021)Jiang, Araki, Ding, and Neubig}]{jiang2021can}
Zhengbao Jiang, Jun Araki, Haibo Ding, and Graham Neubig. 2021.
\newblock How can we know when language models know? on the calibration of language models for question answering.
\newblock \emph{Transactions of the Association for Computational Linguistics}, 9:962--977.

\bibitem[{Kadavath et~al.(2022)Kadavath, Conerly, Askell, Henighan, Drain, Perez, Schiefer, Hatfield-Dodds, DasSarma, Tran-Johnson et~al.}]{kadavath2022language}
Saurav Kadavath, Tom Conerly, Amanda Askell, Tom Henighan, Dawn Drain, Ethan Perez, Nicholas Schiefer, Zac Hatfield-Dodds, Nova DasSarma, Eli Tran-Johnson, et~al. 2022.
\newblock Language models (mostly) know what they know.
\newblock \emph{arXiv preprint arXiv:2207.05221}.

\bibitem[{Kuhn et~al.(2023)Kuhn, Gal, and Farquhar}]{kuhn2023semantic}
Lorenz Kuhn, Yarin Gal, and Sebastian Farquhar. 2023.
\newblock Semantic uncertainty: Linguistic invariances for uncertainty estimation in natural language generation.
\newblock \emph{arXiv preprint arXiv:2302.09664}.

\bibitem[{Kwiatkowski et~al.(2019)Kwiatkowski, Palomaki, Redfield, Collins, Parikh, Alberti, Epstein, Polosukhin, Devlin, Lee et~al.}]{kwiatkowski2019natural}
Tom Kwiatkowski, Jennimaria Palomaki, Olivia Redfield, Michael Collins, Ankur Parikh, Chris Alberti, Danielle Epstein, Illia Polosukhin, Jacob Devlin, Kenton Lee, et~al. 2019.
\newblock Natural questions: a benchmark for question answering research.
\newblock \emph{Transactions of the Association for Computational Linguistics}, 7:453--466.

\bibitem[{Lin et~al.(2022)Lin, Hilton, and Evans}]{lin2022teaching}
Stephanie Lin, Jacob Hilton, and Owain Evans. 2022.
\newblock Teaching models to express their uncertainty in words.
\newblock \emph{arXiv preprint arXiv:2205.14334}.

\bibitem[{Manakul et~al.(2023)Manakul, Liusie, and Gales}]{manakul2023selfcheckgpt}
Potsawee Manakul, Adian Liusie, and Mark~JF Gales. 2023.
\newblock Selfcheckgpt: Zero-resource black-box hallucination detection for generative large language models.
\newblock \emph{arXiv preprint arXiv:2303.08896}.

\bibitem[{Ni et~al.(2024{\natexlab{a}})Ni, Bi, Guo, and Cheng}]{ni2024llms}
Shiyu Ni, Keping Bi, Jiafeng Guo, and Xueqi Cheng. 2024{\natexlab{a}}.
\newblock When do llms need retrieval augmentation? mitigating llms' overconfidence helps retrieval augmentation.
\newblock \emph{arXiv preprint arXiv:2402.11457}.

\bibitem[{Ni et~al.(2024{\natexlab{b}})Ni, Bi, Yu, and Guo}]{ni2024large}
Shiyu Ni, Keping Bi, Lulu Yu, and Jiafeng Guo. 2024{\natexlab{b}}.
\newblock Are large language models more honest in their probabilistic or verbalized confidence?
\newblock \emph{arXiv preprint arXiv:2408.09773}.

\bibitem[{Si et~al.(2022)Si, Gan, Yang, Wang, Wang, Boyd-Graber, and Wang}]{si2022prompting}
Chenglei Si, Zhe Gan, Zhengyuan Yang, Shuohang Wang, Jianfeng Wang, Jordan Boyd-Graber, and Lijuan Wang. 2022.
\newblock Prompting gpt-3 to be reliable.
\newblock \emph{arXiv preprint arXiv:2210.09150}.

\bibitem[{Slobodkin et~al.(2023)Slobodkin, Goldman, Caciularu, Dagan, and Ravfogel}]{slobodkin2023curious}
Aviv Slobodkin, Omer Goldman, Avi Caciularu, Ido Dagan, and Shauli Ravfogel. 2023.
\newblock The curious case of hallucinatory (un) answerability: Finding truths in the hidden states of over-confident large language models.
\newblock In \emph{Proceedings of the 2023 Conference on Empirical Methods in Natural Language Processing}, pages 3607--3625.

\bibitem[{Sprague et~al.(2024)Sprague, Yin, Rodriguez, Jiang, Wadhwa, Singhal, Zhao, Ye, Mahowald, and Durrett}]{sprague2024cot}
Zayne Sprague, Fangcong Yin, Juan~Diego Rodriguez, Dongwei Jiang, Manya Wadhwa, Prasann Singhal, Xinyu Zhao, Xi~Ye, Kyle Mahowald, and Greg Durrett. 2024.
\newblock To cot or not to cot? chain-of-thought helps mainly on math and symbolic reasoning.
\newblock \emph{arXiv preprint arXiv:2409.12183}.

\bibitem[{Su et~al.(2024)Su, Wang, Ai, Hu, Wu, Zhou, and Liu}]{su2024unsupervised}
Weihang Su, Changyue Wang, Qingyao Ai, Yiran Hu, Zhijing Wu, Yujia Zhou, and Yiqun Liu. 2024.
\newblock Unsupervised real-time hallucination detection based on the internal states of large language models.
\newblock \emph{arXiv preprint arXiv:2403.06448}.

\bibitem[{Tian et~al.(2023)Tian, Mitchell, Zhou, Sharma, Rafailov, Yao, Finn, and Manning}]{tian2023just}
Katherine Tian, Eric Mitchell, Allan Zhou, Archit Sharma, Rafael Rafailov, Huaxiu Yao, Chelsea Finn, and Christopher~D Manning. 2023.
\newblock Just ask for calibration: Strategies for eliciting calibrated confidence scores from language models fine-tuned with human feedback.
\newblock \emph{arXiv preprint arXiv:2305.14975}.

\bibitem[{Touvron et~al.(2023)Touvron, Martin, Stone, Albert, Almahairi, Babaei, Bashlykov, Batra, Bhargava, Bhosale et~al.}]{touvron2023llama}
Hugo Touvron, Louis Martin, Kevin Stone, Peter Albert, Amjad Almahairi, Yasmine Babaei, Nikolay Bashlykov, Soumya Batra, Prajjwal Bhargava, Shruti Bhosale, et~al. 2023.
\newblock Llama 2: Open foundation and fine-tuned chat models.
\newblock \emph{arXiv preprint arXiv:2307.09288}.

\bibitem[{Wei et~al.(2022)Wei, Wang, Schuurmans, Bosma, Xia, Chi, Le, Zhou et~al.}]{wei2022chain}
Jason Wei, Xuezhi Wang, Dale Schuurmans, Maarten Bosma, Fei Xia, Ed~Chi, Quoc~V Le, Denny Zhou, et~al. 2022.
\newblock Chain-of-thought prompting elicits reasoning in large language models.
\newblock \emph{Advances in neural information processing systems}, 35:24824--24837.

\bibitem[{Xiong et~al.(2023)Xiong, Hu, Lu, Li, Fu, He, and Hooi}]{xiong2023can}
Miao Xiong, Zhiyuan Hu, Xinyang Lu, Yifei Li, Jie Fu, Junxian He, and Bryan Hooi. 2023.
\newblock Can llms express their uncertainty? an empirical evaluation of confidence elicitation in llms.
\newblock \emph{arXiv preprint arXiv:2306.13063}.

\bibitem[{Yang et~al.(2024)Yang, Yang, Hui, Zheng, Yu, Zhou, Li, Li, Liu, Huang et~al.}]{yang2024qwen2}
An~Yang, Baosong Yang, Binyuan Hui, Bo~Zheng, Bowen Yu, Chang Zhou, Chengpeng Li, Chengyuan Li, Dayiheng Liu, Fei Huang, et~al. 2024.
\newblock Qwen2 technical report.
\newblock \emph{arXiv preprint arXiv:2407.10671}.

\bibitem[{Yang et~al.(2023)Yang, Chern, Qiu, Neubig, and Liu}]{yang2023alignment}
Yuqing Yang, Ethan Chern, Xipeng Qiu, Graham Neubig, and Pengfei Liu. 2023.
\newblock Alignment for honesty.
\newblock \emph{arXiv preprint arXiv:2312.07000}.

\bibitem[{Yang et~al.(2018)Yang, Qi, Zhang, Bengio, Cohen, Salakhutdinov, and Manning}]{yang2018hotpotqa}
Zhilin Yang, Peng Qi, Saizheng Zhang, Yoshua Bengio, William~W Cohen, Ruslan Salakhutdinov, and Christopher~D Manning. 2018.
\newblock Hotpotqa: A dataset for diverse, explainable multi-hop question answering.
\newblock \emph{arXiv preprint arXiv:1809.09600}.

\bibitem[{Yin et~al.(2023)Yin, Sun, Guo, Wu, Qiu, and Huang}]{yin2023large}
Zhangyue Yin, Qiushi Sun, Qipeng Guo, Jiawen Wu, Xipeng Qiu, and Xuanjing Huang. 2023.
\newblock Do large language models know what they don't know?
\newblock \emph{arXiv preprint arXiv:2305.18153}.

\end{thebibliography}
\bibliographystyle{acl_natbib}

\clearpage
\appendix

\section{Appendix}
\label{sec:appendix}

\subsection{Prompts \label{sec:prompts}}
\paragraph{Candidate Answers Generation.} \textit{Generate 10 possible answers for the following question, each separated by a semicolon. These 10 answers must be different, and your response should be as concise as possible, with no irrelevant words beyond the answers.} \\
\textit{Question: [Question]}\\
\textit{Answer:}

\paragraph{Vanilla.} \textit{Answer the following question based on your internal knowledge with one or few words.}\\
\textit{Question: [Question]}\\
\textit{Answer:}

\paragraph{COT.} \textit{Answer the question by briefly explaining your reasoning with one or few sentences, then provide the final answer.}\\
\textit{Question: [Question]} \\
\textit{Answer:}

\paragraph{MC Vanilla.} \textit{The following are multiple choice questions (with answers). Select the correct answer without any irrelevant words. Do not include conversational words and do not provide any explanation.} \\
\textit{Question: [Question]} \\
\textit{Answer:}

\paragraph{MC COT.} \textit{The following are multiple choice questions (with answers){subject}. Briefly explain your reasoning with one or few sentences and choose the correct answer. Start with “So, the correct answer is” to select the correct answer.} \\
\textit{Question: [Question]} \\
\textit{Answer:}

\subsection{Data Selection \label{sec:data selection}}
For the NQ dataset, we use its test set as our NQ-test, the validation set as NQ-dev, and randomly sample 10,000 examples from the training set as NQ-train. For HotpotQA, similar to previous work~\citep{ni2024llms}, we use the validation set as the HQ-test. Additionally, we randomly sample non-overlapping 10,000 and 6,000 examples from the training set as the HQ-train and HQ-dev, respectively. For MMLU, we randomly sample 50\% of its test set as MMLU-train, and split the remaining test set equally into MMLU-dev and MMLU-test. Count of samples for each dataset can be seen in Table~\ref{tab:data_count}.
\begin{table}[h!]
    \centering
    \begin{tabular}{ccccc}
        \toprule
          Dataset & Train & Dev & Test\\
         \midrule
          NQ & 10,000 & 6,489 & 3,610\\
          HotpotQA  &  10,000 & 6,000 & 7,405\\ 
          MMLU & 7,021 & 3,510 & 3,511 \\
         \bottomrule
    \end{tabular}
    \caption{Count of samples for each dataset.}
    \label{tab:data_count}
\end{table}

\begin{figure}[h]
  \centering
    \includegraphics[width=0.45\textwidth]{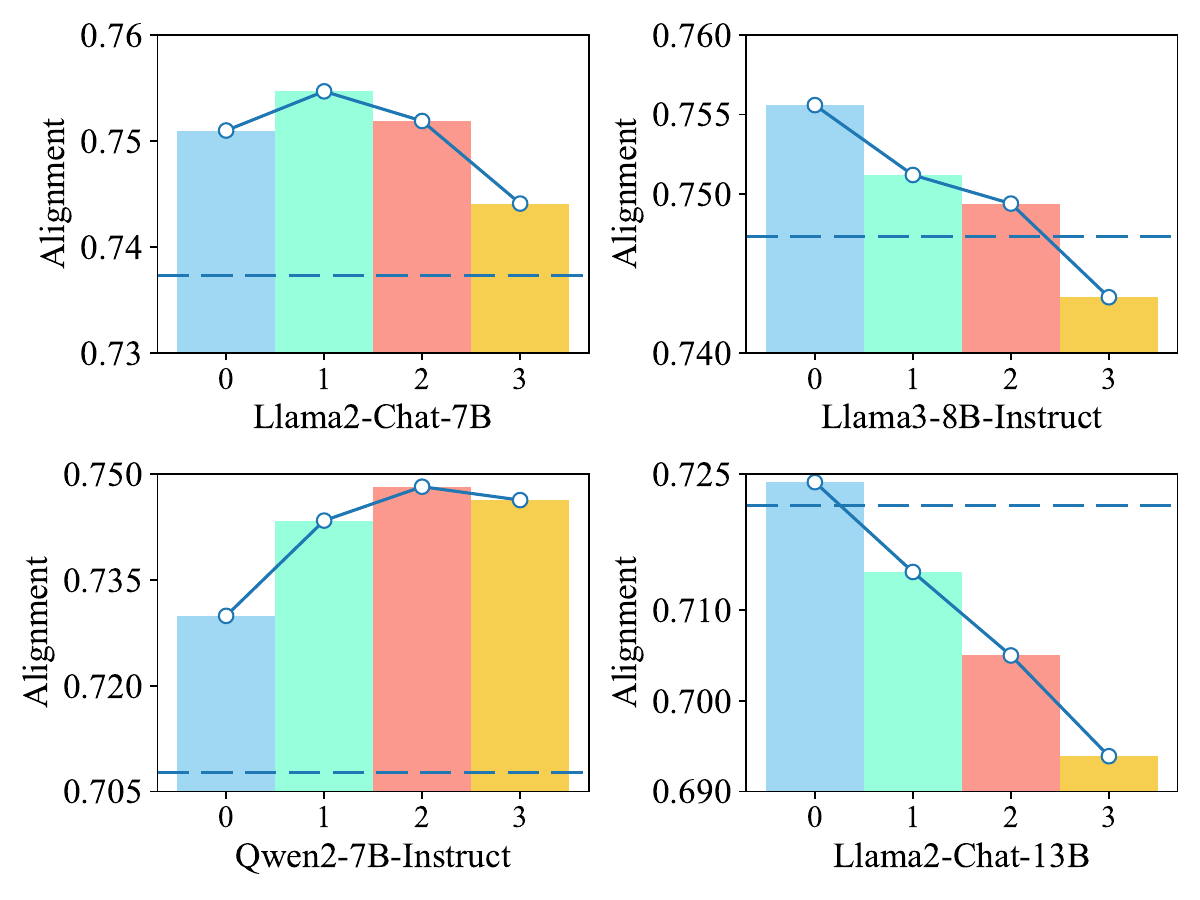}
  \caption{The alignment scores of $C^3$ under different $\beta$ when using Last State on NQ. The horizontal line represents the model's alignment under the vanilla approach.}
  \label{fig:align_nq}
\end{figure}
\subsection{Analysis on Candidate Answers Generation \label{sec:candidate answers analysis}}
\paragraph{Unique Answers Count.}
In this paper, we ask the model to generate 10 candidate answers for each free-form question. The number of remaining answers after deduplication is presented in Table~\ref{tab:choice count}. The table shows that, across all the datasets, all the models generate an average of more than 8 unique answers. Based on this, we reformulate the free-form question into 4 multiple-choice questions, with the count of options for each MC question being 2, 4, 6, and 8, respectively.

\begin{table}[ht]
\centering
\begin{tabular}{ccccc}
\toprule
\textbf{Datasets} & \textbf{Models} & \textbf{Train} & \textbf{Dev} & \textbf{Test} \\
\midrule
\multirow{4}{*}{NQ} & Llama2-7B & 8.50  & 8.51 & 8.29 \\
& Llama3-8B & 9.30  & 9.29 & 9.43 \\
& Qwen2-7B & 9.47 & 9.44 & 9.46 \\
& Llama2-13B & 8.56 & 8.55 & 8.46 \\
\midrule
\multirow{4}{*}{HQ} & Llama2-7B & 8.84 & 8.80  & 8.67 \\
& Llama3-8B & 9.03 & 9.01 & 9.00 \\
& Qwen2-7B & 9.71 & 9.67 & 9.75 \\
& Llama2-13B & 8.70  & 8.65 & 8.53 \\
\bottomrule
\end{tabular}
\caption{The count of unique answers in the generated candidate answers for the NQ and HotpotQA datasets.}
\label{tab:choice count}
\end{table}

\begin{figure*}[h]
  \centering
    \includegraphics[width=\textwidth]{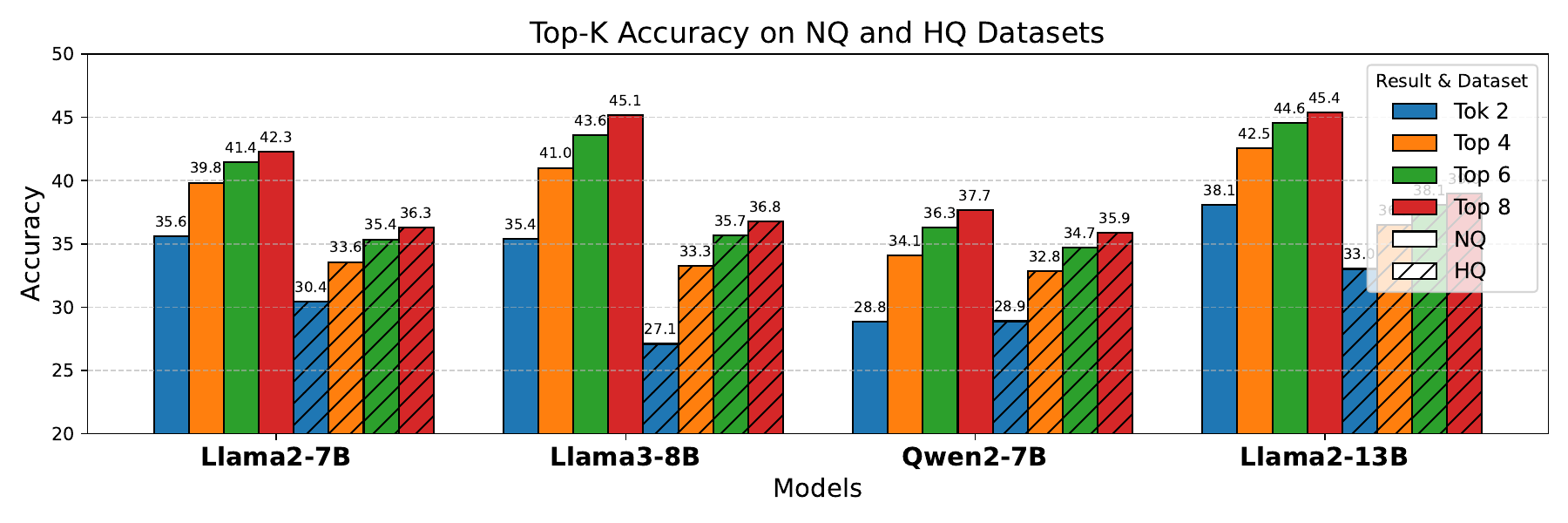}
  \caption{The proportion of the ground-truth answer included in the top-k answers generated by different models on the NQ and HotpotQA test sets.}
  \label{fig:top-k acc test}
\end{figure*}

\begin{table*}[t!]
\centering
\small
\resizebox{\linewidth}{!}{
{\begin{tabular}{cccccccccc}
    \toprule
     \multirow{2}{*}{\textbf{Datasets}} & \multirow{2}{*}{\textbf{Metrics}}  & \multicolumn{2}{c}{\textbf{Llama2-7B}} & \multicolumn{2}{c}{\textbf{Llama3-8B}} & \multicolumn{2}{c}{\textbf{Qwen2-7B}} & \multicolumn{2}{c}{\textbf{Llama2-13B}} \\
    \cmidrule(lr){3-4} \cmidrule(lr){5-6} \cmidrule(lr){7-8} \cmidrule(lr){9-10}
     &   & Vanilla & COT & Vanilla & COT & Vanilla & COT & Vanilla & COT \\
    \midrule
    \multirow{4}{*}{\textbf{NQ}} & Acc & 26.12 & 36.43 & 27.53 & 44.35 & 27.31 & 37.76 & 32.27 & 41.99 \\
    \cmidrule(lr){2-10}
                        & Align-P & 75.38 & 68.40 & 75.57 & 67.43 & 72.73 & 67.50 & 70.13 & 66.28 \\
                        & Align-L & 76.53 & 70.17 & 78.43 & \textbf{70.70} & 75.24 & 70.22 & \textbf{73.92} & 68.13 \\
                        & Align-A & \textbf{77.88} & \textbf{71.19} & \textbf{78.69} & 69.87 & \textbf{76.12} & \textbf{70.94} & 73.72 & \textbf{68.16} \\
    \midrule
    \multirow{4}{*}{\textbf{HotpotQA}} & Acc & 19.93 & 29.55 & 21.63 & 36.79 & 24.96 & 33.34 & 23.69 & 33.10 \\
    \cmidrule(lr){2-10}
                              & Align-P & 80.91 & 76.92 & 82.13 & 74.68 & 79.60 & \textbf{77.25} & 78.89 & 74.71 \\
                              & Align-L & \textbf{82.68} & \textbf{77.41} & 83.29 & \textbf{76.35} & 79.79 & 76.25 & \textbf{80.33} & 75.26 \\
                              & Align-A & 82.59 & 77.07 & \textbf{83.47} & 76.04 & \textbf{79.93} & 77.24 & 79.84 & \textbf{75.28} \\
    \midrule
    \multirow{4}{*}{\textbf{MMLU}} & Acc & 42.20 & 45.51 & 62.49 & 63.77 & 68.72 & 68.63 & 50.58 & 51.18 \\
    \cmidrule(lr){2-10}
                          & Align-P & 65.20 & 65.80 & 71.51 & 69.88 & \textbf{71.08} & 71.43 & 65.51 & 65.41 \\
                          & Align-L & 68.13 & 64.17 & 72.86 & 72.44 & 70.78 & \textbf{73.36} & 67.00 & 66.09 \\
                          & Align-A & \textbf{69.46} & \textbf{68.42} & \textbf{72.98} & \textbf{73.34} & 70.87 & 72.59 & \textbf{68.62} & \textbf{68.15} \\
    \bottomrule
  \end{tabular}}
}
  \caption{QA performance and LLMs' perception of their knowledge boundaries on the NQ, HotpotQA, and MMLU datasets with 10,000 training samples. Bold values indicate the highest performance for each model on each dataset. Align-P, Align-L, and Align-A represent alignment scores using the Pre-generation State, Last State, and Average State, respectively.}
  \label{tab:vanilla and cot res with 10000 training samples}
\end{table*}

\paragraph{Answer Quality.}
We evaluate the proportion of the ground-truth answer included in the top-k answers generated by different models on the NQ and HotpotQA test sets, with results shown in Figure~\ref{fig:top-k acc test}. The figure indicates that as the number of generated answers increases, the top-k accuracy also improves. However, the rate of accuracy growth slows down as the number of answers increases, suggesting that \textbf{LLMs tend to generate correct answers in the earlier positions}. Notably, we do not explicitly instruct the model to prioritize generating the correct answer; it does this autonomously. The proportion of ground-truth answers included in the options is relatively high, which helps prevent situations where the model, despite being confident in its correct answer for the free-form question, becomes uncertain due to the absence of the correct answer among the options. This ensures more accurate calibration and prevents incorrect confidence adjustments.

\begin{table*}[h]
\centering
\scalebox{0.9}{\begin{tabular}{cccccccccc}
    \toprule
     \multirow{2}{*}{\textbf{Models}} & \multirow{2}{*}{\textbf{Methods}} & \multicolumn{4}{c}{\textbf{NQ}} & \multicolumn{4}{c}{\textbf{HotpotQA}} \\
     \cmidrule(lr){3-6} \cmidrule(lr){7-10}
     & & \textbf{Conf.} & \textbf{UPR}$\uparrow$ & \textbf{Overcon.}$\downarrow$  & \textbf{Align.}$\uparrow$ & \textbf{Conf.} &  \textbf{UPR}$\uparrow$ & \textbf{Overcon.}$\downarrow$ & \textbf{Align.}$\uparrow$  \\
    \midrule
    \multirow{2}{*}{\textbf{Llama2-7B}} & Vanilla & 13.50  & 91.50 & 6.28	& 74.82 & 23.43 & 85.26	& 11.80 & 79.91 \\
                       & $C^3$ & 8.86 & \textbf{95.14} & \textbf{3.59}	& \textbf{75.55} & 18.94 	& \textbf{88.56} & \textbf{9.16}  & \textbf{80.70} \\
   \midrule
   \multirow{2}{*}{\textbf{Llama3-8B}} & Vanilla & 19.71 & 88.40 & 8.41	& 75.35 & 24.55 & 85.76 & 11.15 & 80.61 \\
                       & $C^3$ & 15.82 & \textbf{91.31} & \textbf{6.30} & \textbf{75.69} & 18.91 & \textbf{89.79} & \textbf{8.00} & \textbf{81.27} \\
   \midrule
   \multirow{2}{*}{\textbf{Qwen2-7B}}& Vanilla & 26.76	& 81.57 & 13.40 & 72.65 & 27.18 & 83.72	& 12.23 & 77.77 \\
                        & $C^3$& 20.03	& \textbf{87.22} & \textbf{9.29} & \textbf{74.14} & 21.80 & \textbf{87.91}	& \textbf{9.07}  & \textbf{78.69} \\
   \midrule
   \multirow{2}{*}{\textbf{Llama2-13B}} &Vanilla & 29.16 & 81.03 & 12.84 & 71.20	& 44.32 & 75.31 & 12.36	 & 76.32 \\
                       & $C^3$& 17.76 & \textbf{89.91} & \textbf{6.83} & \textbf{71.82} & 22.42 & \textbf{86.29}	& \textbf{10.47} & \textbf{77.80} \\
    \bottomrule
  \end{tabular}}
  \caption{The results of LLMs' perception level of their knowledge boundaries after calibration with $C^3$. Conf., Overcon., and Align. stands for Confident Ratio, Overconfidence, and Alignment, respectively. Bold indicates the best scores for each model and the results are based on Avg State. }
  \label{tab:enhanced align avg}
\end{table*}

\begin{table*}[t!]
\centering
\scalebox{0.8}{\begin{tabular}{cccccccccccc}
    \toprule
    \multirow{2}{*}{Datasets} & \multirow{2}{*}{States} & \multicolumn{5}{c}{Vanilla} & \multicolumn{5}{c}{COT} \\
    \cmidrule(lr){3-7} \cmidrule(lr){8-12}
    &   & Acc & Align.$\uparrow$ & Conf. & Overcon.$\downarrow$ & Conserv.$\downarrow$ & Acc & Align.$\uparrow$ & Conf. & Overcon.$\downarrow$ & Conserv.$\downarrow$ \\
    \midrule
    \multirow{3}{*}{\textbf{NQ}}& Pre-State & 26.1 & 73.65 & 11.09 & \textbf{5.66} & 20.69 & 36.4 & 67.51 & 28.27 & \textbf{12.17} & 20.32 \\
    & Last State & 26.1 & 73.73 & 21.59 & 10.87 & \textbf{15.4} & 36.4 & 68.98 & 32.52 & 13.55 & 17.46 \\
    & Avg State & 26.1 & \textbf{74.82} & 13.5 & 6.28 & 18.9 & 36.4 & \textbf{70.12} & 32.01 & 12.73 & \textbf{17.15} \\
    \midrule
    \multirow{3}{*}{\textbf{HQ}} & Pre-State & 19.9 & 79.69 & 16.47 & \textbf{8.42} & 11.88 & 29.55 & 74.36 & 27.46 & \textbf{11.78} & 13.86 \\
    & Last State & 19.9 & 79.16 & 25.91 & 13.41 & \textbf{7.43} & 29.55 & \textbf{74.77} & 28.74 & 12.21 & 13.02 \\
    & Avg State & 19.9 & \textbf{79.91} & 23.43 & 11.8 & 8.3 & 29.55 & 72.71 & 34.83 & 16.29 & \textbf{11.00} \\
    \midrule
    \multirow{3}{*}{\textbf{MMLU}}& Pre-State & 42.2 & 62.86 & 33.7 & 14.06 & 23.08 & 45.51 & 63.83 & 37.21 & 14.26 & 21.91 \\
    & Last State & 42.2 & 68.11 & 31.4 & \textbf{10.28} & 21.61 & 45.51 & 66.55 & 38.36 & 13.47 & \textbf{19.98} \\
    & Avg State & 42.2 & \textbf{68.71} & 32.13 & 10.35 & \textbf{20.94} & 45.51 & \textbf{67.95} & 34.04 & \textbf{10.61} & 21.44 \\
    \bottomrule
  \end{tabular}}
  \caption{Detailed perception for Llama2-Chat-7B. Align., Conf., Overcon., and Conserv. stands for Alignment, Confidence level, Overconfidence, and Conservativeness, respectively. Bold denotes the best scores on each dataset. }
  \label{tab:detail conf llama7b}
\end{table*}
\subsection{LLMs' QA Performance and Perception on Reformatted Questions}
LLMs' QA performance and perception levels on reformatted questions can be seen in Table~\ref{tab:question reformulation performance}. 

1) \textbf{LLMs can be misled by self-generated answers, leading to worse QA performance.} As the number of options increases, despite the higher likelihood of including the ground-truth answer among them (See Figure~\ref{fig:top-k acc test}), LLMs' QA performance consistently decline. This suggests that the more options there are, the harder it becomes for LLMs. This indicates that the LLMs struggle to select the correct answer when faced with similar self-generated answers.

2) \textbf{LLMs show better perception level on reformatted questions.} The decline in LLMs' confidence on reformatted questions is often less than the decrease in QA performance, which reduces conservativeness and enhances alignment. More accurate assessments of these questions enable us to obtain reliable supplementary information.

3) Transforming a free-form question into a multiple-choice question may improve the model's perception level, but it comes at the cost of QA performance. In contrast, $C^3$ achieves the lowest overconfidence in most cases and consistently enhances alignment without negatively impacting QA performance.




\subsection{The Impact of Training Sample Amount \label{sec:effects of training sample amount}}

The QA performance and alignment results for all the models, trained on 10,000 examples, can be found in Table~\ref{tab:vanilla and cot res with 10000 training samples}. We observe that:

1) The alignment scores can be improved (most are in the 70s, with some even exceeding 80) with a little more training data.

2) Compared to Table~\ref{tab:vanilla and cot res}, the gap between LLMs' perception levels before and after response generation remains nearly unchanged with different training amount.

In this part of the comparison, our focus is not on determining the optimal number of training samples for achieving the best perception level. Instead, we are solely concerned with the gap in the model's perception of its knowledge boundaries before and after generating the answers and the effects of training sample amount.

\begin{figure*}[h]
  \centering
    \includegraphics[width=\textwidth]{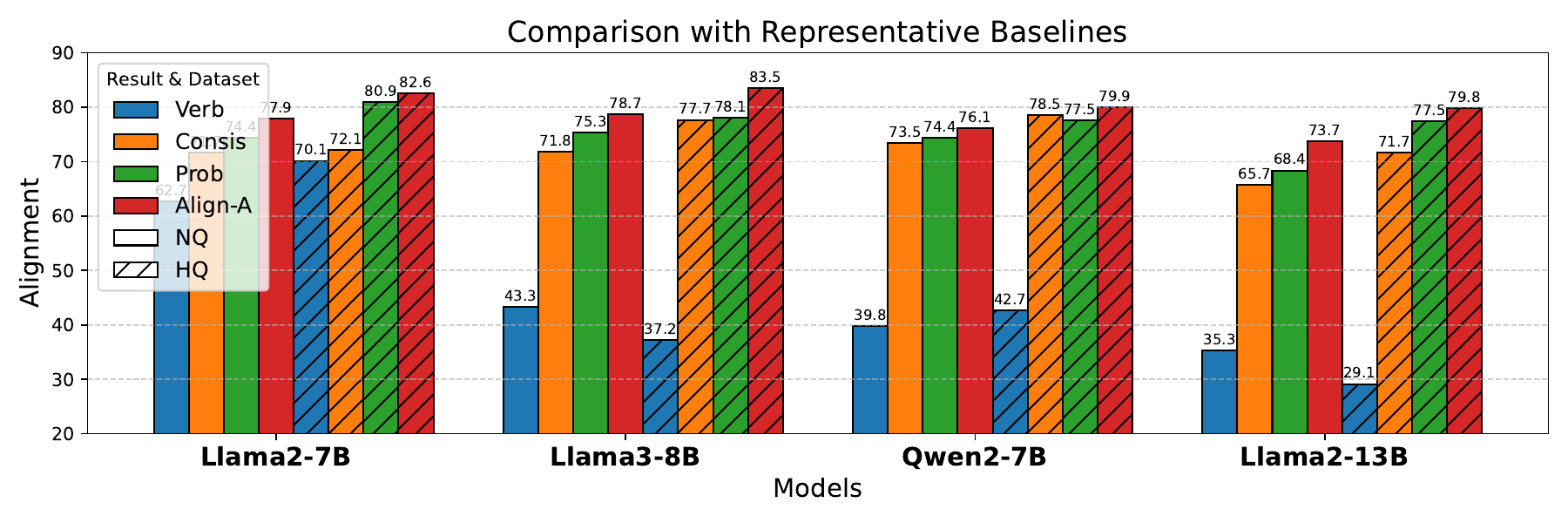}
  \caption{Comparison between Align-A with representative baselines.}
  \label{fig:baselines}
\end{figure*}
\subsection{Alignment of Representative Baselines}

In this section, we introduce several mainstream confidence estimation methods and compare their alignment scores with those obtained from LLMs’ internal states. The methods are:
\begin{itemize}[leftmargin=*,nosep]
    \item Probabilistic Confidence (Prob): This method calculates the average generation probability of all tokens in the response as the model's confidence in the answer~\citep{guo2017calibration}.
    \item Verbalized Confidence (Verb)~\citep{xiong2023can}: Some studies have shown that LLMs can express their confidence in answers using words.
    \item Semantic Consistency-Based Confidence (Consis)~\citep{manakul2023selfcheckgpt}: Rather than relying on probabilistic confidence, this method measures consistency across multiple responses to the same question, using NLI models or LLMs to check if the responses convey the same meaning.
\end{itemize}

\noindent The Vanilla method in our paper is a strong baseline because prior research~\citep{ni2024large, su2024unsupervised} has shown that confidence derived from the model’s internal states (the Vanilla method) more accurately reflects the model's true capabilities compared to other methods like probabilistic confidence, verbalized confidence, and semantic consistency-based confidence. We evaluate these baselines using the following experimental setup:
\begin{itemize}[leftmargin=*,nosep]
\item Answer and confidence generation: For Prob and Verb, we obtain the greedy answer. For Consis, we generate the greedy answer and additionally sample 10 responses with a temperature of 0.8. We use Qwen2.5-32B-Instruct~\citep{yang2024qwen2} to evaluate the semantic consistency between each sampled response and the greedy answer.
\item Confidence binarization: We select the optimal threshold using the training (10,000 samples for each dataset) and development sets to binarize the confidence scores.
\end{itemize}

Figure~\ref{fig:baselines} presents a comparison of the alignment between answer correctness and the confidence estimated by various methods, alongside Align-A. Align-A is trained on the same training set as the baselines (10,000 samples). The results demonstrate that Align-A consistently outperforms all baseline methods across the board. Therefore, in $C^3$, we focus exclusively on further improving performance using hidden states.

\subsection{Costs of $C^3$}
Cost of $C^3$: 1) Cost of question reformulation: This requires one LLM inference call. In this step, the model generates multiple candidate answers in a single generation, and automated rules are used to construct multi-choice questions from these answers. 2) Cost of consistency checking: This requires $\gamma=4$ (i.e., count of constructed multi-choice questions) LLM inference calls because the model is asked to answer four different multiple-choice questions (i.e., with $k=2,4,6,8$ options). We then obtain the model's confidence via the MLP and use an automated program to check the consistency of the confidence. Therefore, $C^3$ requires only $\gamma + 1=5$ LLM inference calls.

Cost of existing semantic consistency-based methods: Semantic consistency-based methods often require generating multiple responses and using powerful LLMs to assess the semantic consistency between texts. In our experiment, generating answers takes 11 inference calls (1 greedy answer + 10 sampling answers), and consistency checking takes 10, totaling 21 inference calls.
Compared to existing semantic consistency-based confidence estimation methods that rely on text semantic consistency, $C^3$ achieves a much lower cost.

\begin{table*}[t!]
\centering
\scalebox{0.72}{
\begin{tabular}{cccccccccccccccc}
    \toprule
    \multirow{2}{*}{\textbf{Models}} & \multirow{2}{*}{\textbf{Methods}} & \multicolumn{5}{c}{\textbf{NQ}} & \multicolumn{5}{c}{\textbf{HotpotQA}} \\
    \cmidrule(lr){3-7} \cmidrule(lr){8-12}
    & & \textbf{Acc} & \textbf{Align.}$\uparrow$ & \textbf{Conf.} & \textbf{Overconf.}$\downarrow$ & \textbf{Conserv.}$\downarrow$ & \textbf{Acc} & \textbf{Align.}$\uparrow$ & \textbf{Conf.} & \textbf{Overconf.}$\downarrow$ & \textbf{Conserv.}$\downarrow$ \\
    \midrule
    \multirow{5}{*}{\textbf{Llama2-7B}} & Vanilla & 26.12 & 74.82 & 13.50 & 6.28 & 18.90 & 19.93 & 79.91 & 23.43 & 11.80 & 8.30 \\
    & MC-2 & 21.91 & 76.17 & 17.73 & 9.82 & 14.01 & 17.58 & 80.31 & 18.51 & 10.31 & 9.38 \\
    & MC4 & 19.09 & 78.04 & 14.76 & 8.82 & 13.14 & 16.34 & 77.79 & 25.80 & 15.84 & \textbf{6.38} \\
    & MC-6 & 15.46 & 79.64 & 20.45 & 12.68 & 7.68 & 13.80 & \textbf{84.17} & 12.04 & \textbf{7.03} & 8.80 \\
    & MC-8 & 13.66 & \textbf{83.20} & 14.80 & 8.98 & \textbf{7.83} & 12.72 & 82.69 & 16.22 & 10.40 & 6.90 \\
    & $C^3$ & 26.12 & 75.55 & 8.86 & \textbf{3.59} & 20.86 & 19.93 & 80.70 & 18.94 & 9.16 & 10.15 \\
    \midrule
    \multirow{5}{*}{\textbf{Llama3-8B}} & Vanilla & 27.53 & 75.35 & 19.71 & 8.41 & 16.23 & 21.63 & \textbf{80.61} & 24.55 & 11.15 & 8.24 \\
    & MC-2 & 26.20 & 75.90 & 26.15 & 12.02 & 12.08 & 19.81 & 80.43 & 19.93 & 9.84 & 9.72 \\
    & MC-4 & 23.07 & 78.46 & 18.60 & 8.53 & 13.01 & 18.65 & 80.40 & 23.92 & 12.44 & 7.17 \\
    & MC-6 & 22.22 & \textbf{79.76} & 17.80 & 7.91 & 12.33 & 17.80 & 80.50 & 23.75 & 12.73 & 6.78 \\
    & MC-8 & 21.0 & 79.09 & 23.93 & 11.92 & \textbf{8.98} & 16.29 & 80.26 & 23.85 & 13.65 & \textbf{6.09} \\
    & $C^3$ & 27.53 & 75.69 & 15.82 & \textbf{6.30} & 18.01 & 21.63 & 81.27 & 18.91 & \textbf{8.00} & 10.73 \\
    \midrule
    \multirow{5}{*}{\textbf{Qwen2-7B}}  & Vanilla & 27.31 & 72.57 & 26.76 & 13.40 & 13.95 & 24.96 & 77.79 & 27.18 & 12.23 & 10.00 \\
    & MC-2 & 21.30 & 75.02 & 26.57 & 15.12 & 9.85 & 20.51 & 80.40 & 21.91 & 10.50 & 9.10 \\
    & MC-4 & 20.08 & \textbf{77.94} & 22.40 & 12.19 & 9.87 & 16.85 & 80.65 & 20.42 & 11.46 & 7.89 \\
    & MC-6 & 17.84 & 77.61 & 22.21 & 13.38 & 9.01 & 15.65 & 81.22 & 18.60 & 10.87 & 7.91 \\
    & MC-8 & 18.12 & 77.42 & 22.77 & 13.62 & \textbf{8.97} & 14.75 & \textbf{81.89} & 17.34 & 10.35 & \textbf{7.76} \\
    & $C^3$ & 27.31 & 74.14 & 20.03 & \textbf{9.29} & 16.57 & 24.96 & 78.69 & 21.80 & \textbf{9.07} & 12.23 \\
    \midrule
    \multirow{5}{*}{\textbf{Llama2-13B}}& Vanilla & 32.27 & 71.20 & 29.16 & 12.84 & 15.96 & 23.69 & 76.32 & 27.27 & 13.63 & 10.05 \\
    & MC-2 & 25.90 & 74.25 & 25.08 & 12.47 & 13.29 & 21.38 & 77.58 & 24.75 & 12.90 & 9.52 \\
    & MC-4 & 22.74 & 76.21 & 21.52 & 11.28 & 12.50 & 18.83 & 78.52 & 24.33 & 13.49 & 7.99 \\
    & MC-6 & 19.31 & 78.64 & 19.48 & 10.77 & 10.59 & 16.87 & 79.39 & 22.29 & 13.01 & 7.59 \\
    & MC-8 & 16.95 & \textbf{80.49} & 18.05 & 10.30 & \textbf{9.21} & 14.79 & \textbf{79.59} & 24.29 & 14.96 & 
    \textbf{5.46} \\
    & $C^3$ & 32.27 & 71.82 & 17.76 & \textbf{6.83} & 21.35 & 23.69 & 77.80 & 22.42 & \textbf{10.47} & 11.74 \\
    \bottomrule
\end{tabular}
}
\caption{LLMs' QA performance and perception level on reformatted MC questions when using Avg State.}
\label{tab:question reformulation performance}
\end{table*}

\begin{table*}[h]
\centering
\scalebox{0.8}{\begin{tabular}{cccccccccccc}
    \toprule
    \multirow{2}{*}{Datasets} & \multirow{2}{*}{States} & \multicolumn{5}{c}{Vanilla} & \multicolumn{5}{c}{COT} \\
    \cmidrule(lr){3-7} \cmidrule(lr){8-12}
    &   & Acc & Align.$\uparrow$ & Conf. & Overcon.$\downarrow$ & Conserv.$\downarrow$ & Acc & Align.$\uparrow$ & Conf. & Overcon.$\downarrow$ & Conserv.$\downarrow$ \\
    \midrule
    \multirow{3}{*}{\textbf{NQ}} & Pre-State & 27.31 & \textbf{72.69} & 0 & \textbf{0} & 27.31 & 37.76 & 64.06 & 29.30 & \textbf{13.74} & 22.2 \\
    & Last State & 27.31 & 70.77 & 29.41 & 15.66 & \textbf{13.56} & 37.76 & 67.85 & 34.27 & 14.33 & 17.82 \\
    & Avg State & 27.31 & 72.65 & 26.76 & 13.4 & 13.95 & 37.76 & \textbf{69.78} & 37.67 & 15.07 & \textbf{15.15} \\
    \midrule
    \multirow{3}{*}{\textbf{HQ}} & Pre-State & 24.96 & \textbf{79.34} & 19.90 & \textbf{7.81} & 12.86 & 33.34 & \textbf{76.79} & 25.59 & \textbf{7.73} & 15.49 \\
    & Last State & 24.96 & 75.13 & 29.95 & 14.93 & \textbf{9.93} & 33.34 & 75.20 & 29.12 & 10.29 & 14.51 \\
    & Avg State & 24.96 & 77.77 & 27.18 & 12.23 & 10.00 & 33.34 & 75.33 & 35.33 & 13.33 & \textbf{11.34} \\
    \midrule
    \multirow{3}{*}{\textbf{MMLU}} & Pre-State & 68.72 & 69.33 & 92.26 & 27.19 & \textbf{3.48} & 68.63 & 68.68 & 92.45 & 27.75 & \textbf{3.57} \\
    & Last State & 68.72 & 70.02 & 87.58 & 24.50 & 5.48 & 68.63 & 72.66 & 81.46 & \textbf{20.26} & 7.07 \\
    & Avg State & 68.72 & \textbf{72.57} & 82.27 & \textbf{20.57} & 6.85 & 68.63 & \textbf{72.74} & 85.76 & 22.38 & 4.89 \\
    \bottomrule
  \end{tabular}}
  \caption{Detailed perception for Qwen2-7B-Instruct. Align., Conf., Overcon., and Conserv. stands for Alignment, Confidence level, Overconfidence, and Conservativeness, respectively. Bold denotes the best scores on each dataset. Conf=0 is due to insufficient training on the pre-state. Training on 10,000 data can address this.}
  \label{tab:detail conf qwen2}
\end{table*}

\begin{table*}[h]
\centering
\scalebox{0.8}{\begin{tabular}{cccccccccccc}
    \toprule
    \multirow{2}{*}{Datasets} & \multirow{2}{*}{States} & \multicolumn{5}{c}{Vanilla} & \multicolumn{5}{c}{COT} \\
    \cmidrule(lr){3-7} \cmidrule(lr){8-12}
    &   & Acc & Align.$\uparrow$ & Conf. & Overcon.$\downarrow$ & Conserv.$\downarrow$ & Acc & Align.$\uparrow$ & Conf. & Overcon.$\downarrow$ & Conserv.$\downarrow$ \\
    \midrule
    \multirow{3}{*}{\textbf{NQ}} & Pre-State & 32.27 & 68.67 & 18.91 & \textbf{8.98} & 22.35 & 41.99 & 65.16 & 24.72 & \textbf{8.78} & 26.06 \\
    & Last State & 32.27 & \textbf{72.15} & 26.92 & 11.25 & 16.60 & 41.99 & 66.12 & 35.47 & 13.67 & 20.20 \\
    & Avg State & 32.27 & 71.20 & 29.16 & 12.84 & \textbf{15.96} & 41.99 & \textbf{67.03} & 35.60 & 13.29 & \textbf{19.69} \\
    \midrule
    \multirow{3}{*}{\textbf{HQ}} & Pre-State & 23.69 & 75.91 & 24.26 & 12.33 & 11.76 & 33.1 & 73.55 & 24.11 & \textbf{8.73} & 17.72 \\
    & Last State & 23.69 & \textbf{77.66} & 25.10 & \textbf{11.88} & 10.46 & 33.10 & \textbf{74.10} & 28.76 & 10.78 & 15.12 \\
    & Avg State & 23.69 & 76.32 & 27.27 & 13.63 & \textbf{10.05} & 33.10 & 72.57 & 30.79 & 12.56 & \textbf{14.87} \\
    \midrule
    \multirow{3}{*}{\textbf{MMLU}} & Pre-State & 50.58 & 64.88 & 55.23 & 19.97 & \textbf{15.15} & 51.18 & 64.25 & 49.76 & 17.77 & 17.97 \\
    & Last State & 50.58 & 67.75 & 51.58 & 16.71 & 15.54 & 51.18 & 67.96 & 54.94 & 18.51 & \textbf{13.53} \\
    & Avg State & 50.58 & \textbf{69.18} & 44.32 & \textbf{12.36} & 18.46 & 51.18 & \textbf{69.30} & 46.58 & \textbf{13.66} & 17.04 \\
    \bottomrule
  \end{tabular}}
  \caption{Detailed perception for Llama2-Chat-13B. Align., Conf., Overcon., and Conserv. stands for Alignment, Confidence level, Overconfidence, and Conservativeness, respectively. Bold denotes the best scores on each dataset. }
  \label{tab:detail conf llama13b}
\end{table*}

\end{document}